\documentclass{article} 
\usepackage{iclr2023_conference,times}

\usepackage{amsmath,amsfonts,bm}

\def\eqref#1{equation~\ref{#1}}

\def\1{\bm{1}}

\DeclareMathAlphabet{\mathsfit}{\encodingdefault}{\sfdefault}{m}{sl}
\SetMathAlphabet{\mathsfit}{bold}{\encodingdefault}{\sfdefault}{bx}{n}

\usepackage{subcaption}
\usepackage{booktabs}
\usepackage{csquotes}
\usepackage{graphicx}
\usepackage{mathtools}
\usepackage{multirow}
\usepackage{hyperref}
\usepackage{verbatim}
\usepackage{wrapfig}
\usepackage{float}
\usepackage{url}

\renewcommand{\eqref}[1]{Eq. (\ref{#1})}
\newcommand{\s}{{\boldsymbol s}}

\newcommand{\epsilonb}{{\boldsymbol \epsilon}}

\title{Denoising MCMC for Accelerating Diffusion-Based Generative Models}

\author{Beomsu Kim \\
Dept. of Mathematical Sciences, KAIST \\
\texttt{beomsu.kim@kaist.ac.kr} \\
\And
Jong Chul Ye \\
Graduate School of AI, KAIST \\
\texttt{jong.ye@kaist.ac.kr}
}

\newcommand{\NN}{\mathcal{N}}
\newcommand{\EE}{\mathbb{E}}
\newcommand{\RR}{\mathbb{R}}
\newcommand{\XX}{\mathcal{X}}
\renewcommand{\SS}{\mathcal{S}}

\iclrfinalcopy 
\begin{document}

\maketitle

\begin{abstract}
Diffusion models are powerful generative models that simulate the reverse of diffusion processes using score functions to synthesize data from noise. The sampling process of diffusion models can be interpreted as solving the reverse stochastic differential equation (SDE) or the ordinary differential equation (ODE) of the diffusion process, which often requires up to thousands of discretization steps to generate a single image. This has sparked a great interest in developing efficient integration techniques for reverse-S/ODEs. Here, we propose an orthogonal approach to accelerating score-based sampling: Denoising MCMC (DMCMC). DMCMC first uses MCMC to produce samples in the product space of data and variance (or diffusion time). Then, a reverse-S/ODE integrator is used to denoise the MCMC samples. Since MCMC traverses close to the data manifold, the computation cost of producing a clean sample for DMCMC is much less than that of producing a clean sample from noise. To verify the proposed concept, we show that Denoising Langevin Gibbs (DLG), an instance of DMCMC, successfully accelerates all six reverse-S/ODE integrators considered in this work on the tasks of CIFAR10 and CelebA-HQ-256 image generation. Notably, combined with integrators of \citet{karras2022} and pre-trained score models of \citet{song2021}, DLG achieves SOTA results. In the limited number of score function evaluation (NFE) settings on CIFAR10, we have $3.86$ FID with $\approx 10$ NFE and $2.63$ FID with $\approx 20$ NFE. On CelebA-HQ-256, we have $6.99$ FID with $\approx 160$ NFE, which beats the current best record of \citet{kim2022} among score-based models, $7.16$ FID with $4000$ NFE.
Code: \url{https://github.com/1202kbs/DMCMC}
\end{abstract}

\section{Introduction}

Sampling from a probability distribution given its score function, i.e., the gradient of the log-density, is an active area of research in machine learning. Its applications range far and wide, from Bayesian learning \citep{welling2011} to learning energy-based models \citep{song2021ebm}, synthesizing new high-quality data \citep{dhariwal2021}, and so on. Typical examples of traditional score-based samplers are Markov chain Monte Carlo (MCMC) methods such as Langevin dynamics \citep{langevin1908} and Hamiltonian Monte Carlo \citep{neal2011}.

Recent developments in score matching with deep neural networks (DNNs) have made it possible to estimate scores of high-dimensional distributions such as those of natural images \citep{song2019score}. However, natural data distributions are often sharp and multi-modal, rendering na\"{i}ve application of traditional MCMC methods impractical. Specifically, MCMC methods tend to skip over or get stuck at local high-density modes, producing biased samples \citep{levy2018}.

Diffusion models \citep{dickstein2015,ho2020,song2021ddim} depart from MCMC and use the concept of diffusion, the process of gradually corrupting data into noise, to generate samples. \citet{song2021} observed that for each diffusion process, there is a reverse stochastic differential equation (SDE) and an ordinary differential equation (ODE). Hence, given a noise sample, integrating the reverse-S/ODE produces a data sample. Only a time-dependent score function of the data during the diffusion process is required to simulate the reverse process.

This discovery generated great interest in finding better ways to integrate reverse-S/ODEs. For instance, \citet{song2021} uses black-box ODE solvers with adaptive stepsizes to accelerate sampling. Furthermore, multitude of recent works on score-based generative modeling focus on improving reverse-S/ODE integrators \citep{martineau2021,lu2022,karras2022,zhang2022}.

\begin{figure}[t]
\centering
\begin{subfigure}{0.9\linewidth}
\includegraphics[width=1.0\linewidth]{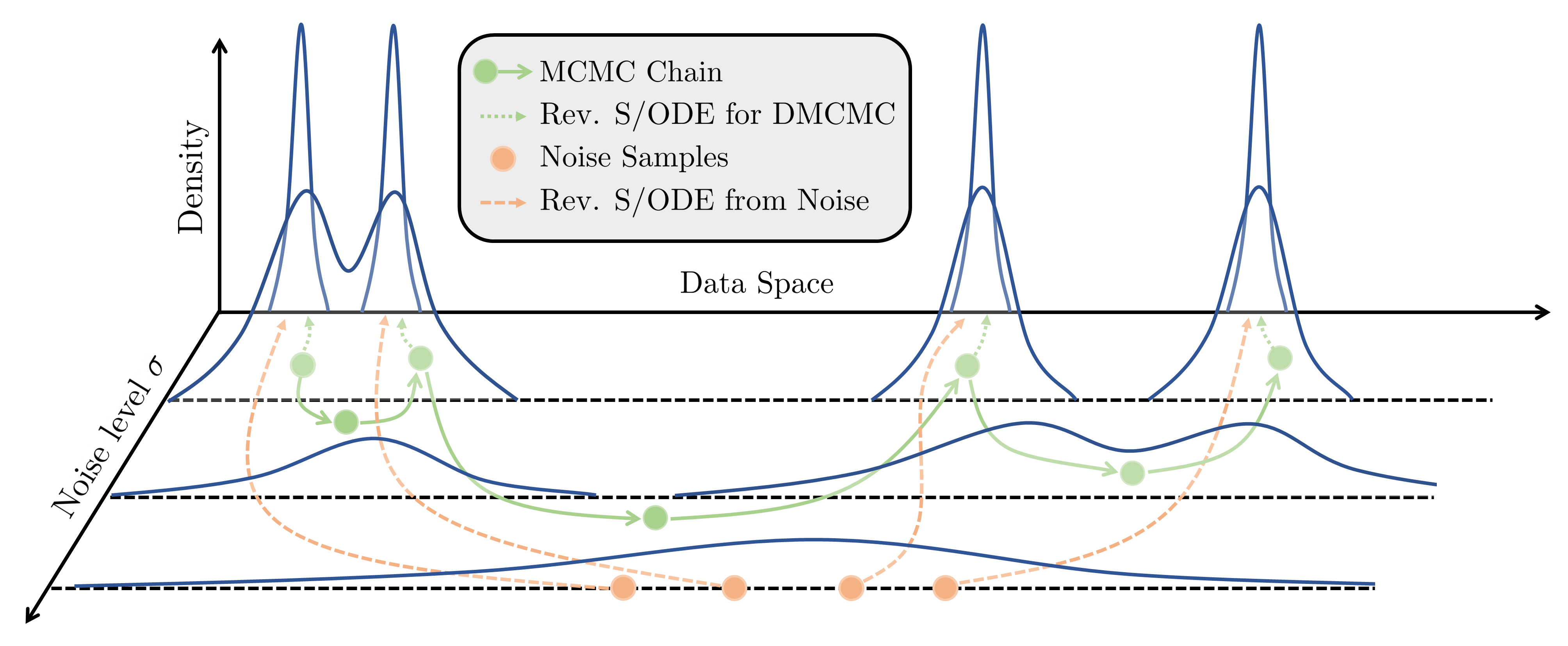}
\end{subfigure}
\begin{subfigure}{0.49\linewidth}
\includegraphics[width=1.0\linewidth]{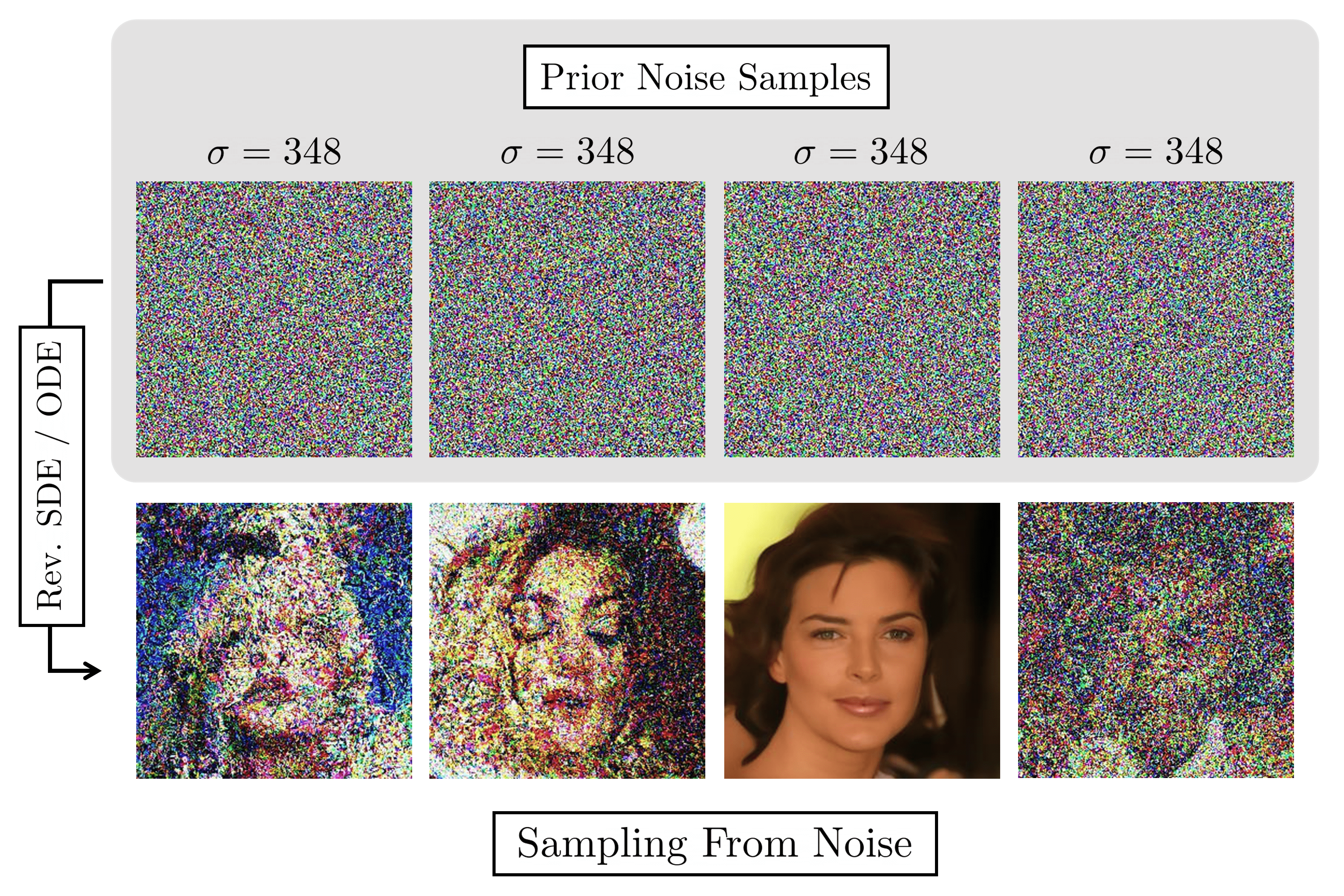}
\end{subfigure}
\hfill
\begin{subfigure}{0.49\linewidth}
\includegraphics[width=1.0\linewidth]{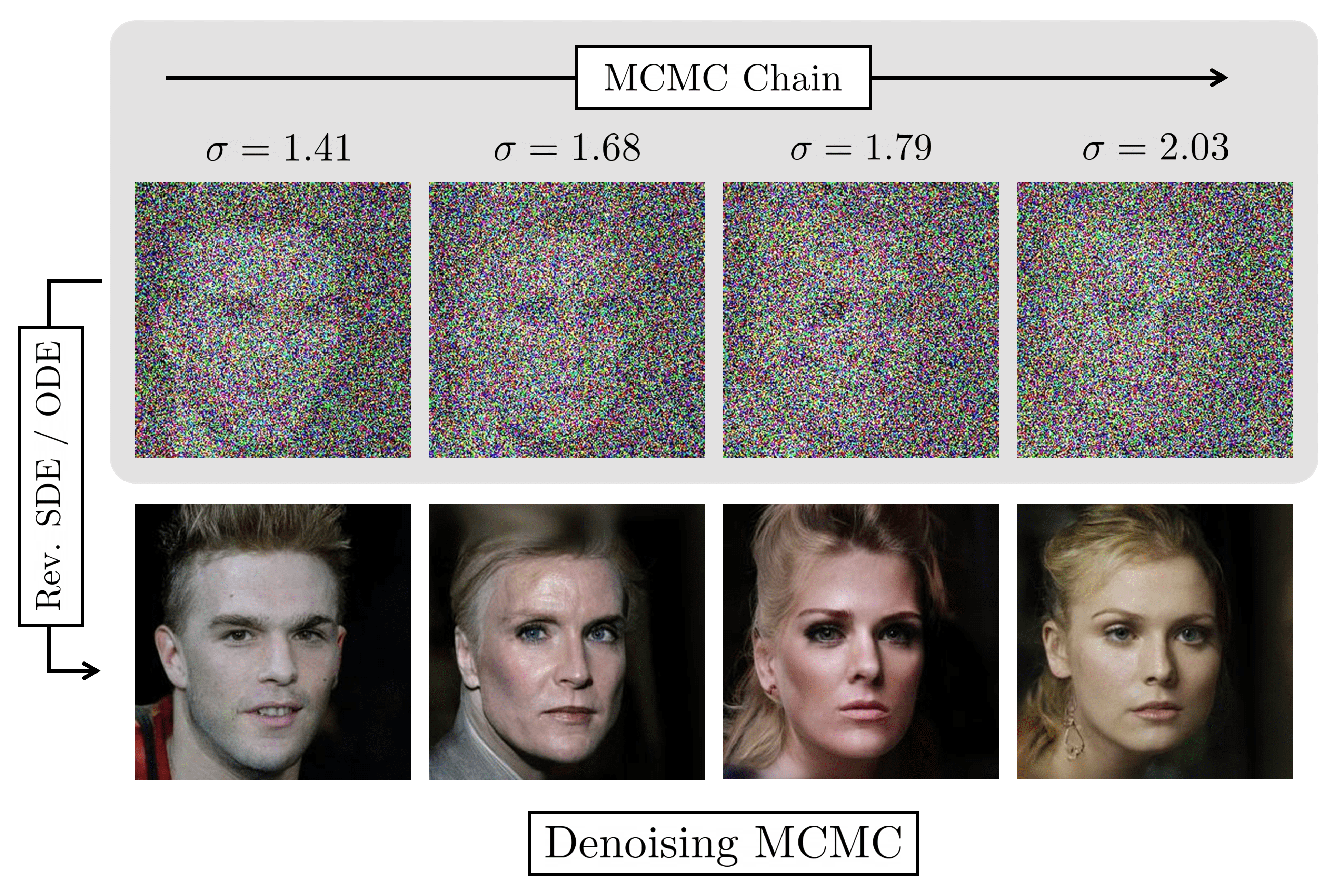}
\end{subfigure}
\caption{\textbf{Top:} a conceptual illustration of a VE diffusion model sampling process and DMCMC sampling process. VE diffusion models integrate the reverse-S/ODE starting from maximum diffusion time / maximum noise level. So, samples are often noisy with small computation budget due to large truncation error. DMCMC produces an MCMC chain which travels close to the image manifold (compare the noise level $\sigma$). So, the MCMC samples can be denoised to produce high-quality data with relatively little computation. \textbf{Bottom:}
Visualization of sampling processes without (left) and with (right) DMCMC on CelebA-HQ-256 under a fixed computation budget.}
\label{fig:dmcmc_example}
\end{figure}

In this work, we develop an orthogonal approach to accelerating score-based sampling. Specifically, we propose Denoising MCMC (DMCMC) which combines MCMC with reverse-S/ODE integrators. MCMC is used to generate samples $\{(\bm{x}_n,t_n)\}$ in the product space of data $\bm{x}$ and variance exploding (VE) diffusion time $t$ / noise level $\sigma$ (see Fig.~\ref{fig:dmcmc_example} top panel). Since all modes are connected in the product space, MCMC mixes well. Then, a reverse-S/ODE integrator solves the reverse-S/ODE starting at $\bm{x}_n$ from time $t = t_n$ to $t = 0$. Since MCMC explores high-density regions, the MCMC chain stays close to the data manifold, so $t_n$ tends to be close to $0$, i.e., noise level tends to be small (see Fig.~\ref{fig:dmcmc_example} top and bottom panels). Thus, integrating the reverse-S/ODE from $t = t_n$ to $t = 0$ is much faster than integrating the reverse-S/ODE from maximum time $t = T$ to $t = 0$ starting from noise. This leads to a significant acceleration of the sampling process.

Our contributions can be summarized as follows.
\begin{itemize}
\item We introduce the product space of data and diffusion time, and develop a novel score-based sampling framework called Denoising MCMC on the product space. Our framework is general, as any MCMC, any VE process noise-conditional score function, and any reverse-S/ODE integrator can be used in a plug-and-play manner.
\item We develop Denoising Langevin Gibbs (DLG), which is an instance of Denoising MCMC that is simple to implement and is scalable. The MCMC part of DLG alternates between a data update step with Langevin dynamics and a noise level prediction step, so all that DLG requires is a pre-trained noise-conditional score network and a noise level classifier.
\item We verify the effectiveness of DLG by accelerating six reverse-S/ODE integrators. Notably, combined with the integrators of \citet{karras2022}, DLG achieves state-of-the-art results. On CIFAR10 in the limited number of score function evaluation (NFE) setting, we obtain $3.86$ FID with $\approx 10$ NFE and $2.63$ FID with $\approx 20$ NFE. On CelebA-HQ-256, we have $6.99$ FID with $\approx 160$ NFE, which is currently the best result with score-based models. The computation cost of evaluating a noise level classifier is negligible, so we obtain acceleration essentially for free.
\end{itemize}

\section{Background}

\subsection{Denoising Score Matching}

Given a distribution $p(\bm{x})$, a noise level $\sigma$, and a perturbation kernel $p_\sigma(\bm{x}\mid\tilde{\bm{x}}) = \NN(\bm{x} \mid \tilde{\bm{x}}, \sigma^2 \bm{I})$, solving the denoising score matching objective \citep{vincent2011}
\begin{align}
\min_\theta \EE_{p(\tilde{\bm{x}})} \EE_{p_\sigma(\bm{x} \mid \tilde{\bm{x}})} \left[ \| \s_\theta(\bm{x}) - \nabla_{\bm{x}} \log p_\sigma(\bm{x} \mid \tilde{\bm{x}}) \|_2^2 \right] \label{eq:dsm_orig}
\end{align}
yields a score model $\s_\theta(\bm{x})$ which approximates the score of $\int p_\sigma(\bm{x} \mid \tilde{\bm{x}}) p(\tilde{\bm{x}}) \, d\tilde{\bm{x}}$.
Denoising score matching was then extended to train Noise Conditional Score Networks (NCSNs) $\s_\theta(\bm{x}, \sigma)$ which approximate the score of data smoothed at a general set of noise levels by solving
\begin{align}
\min_\theta \EE_{\lambda(\sigma)} \EE_{p(\tilde{\bm{x}})} \EE_{p_{\sigma}(\bm{x} \mid \tilde{\bm{x}})} \left[ \| \s_\theta(\bm{x},\sigma) - \nabla_{\bm{x}} \log p_{\sigma}(\bm{x} \mid \tilde{\bm{x}}) \|_2^2 \right] \label{eq:dsm_cont}
\end{align}
where $\lambda(\sigma)$ can be a discrete or a continuous distribution over $(\sigma_{\min},\sigma_{\max})$ \citep{song2019score,song2021}. We note $\int p_\sigma(\bm{x} \mid \tilde{\bm{x}}) p(\tilde{\bm{x}}) \, d\tilde{\bm{x}}$ approaches $p(\bm{x})$ as $\sigma \rightarrow 0$, since the perturbation kernel $p_\sigma(\bm{x} \mid \tilde{\bm{x}})$ converges to the Dirac delta function centered at $\bm{x}$.

\subsection{Markov Chain Monte Carlo (MCMC)}

Given an unnormalized version of $p(\bm{x})$ or the score function $\nabla_{\bm{x}} \log p(\bm{x})$, MCMC constructs a Markov chain in the data space whose stationary distribution is $p(\bm{x})$. An MCMC which uses the unnormalized density is the Metropolis-Hastings MCMC \citep{metropolis1953,hastings1970} that builds a Markov chain by sequentially accepting or rejecting proposal distribution samples according to a density ratio. A popular score-based MCMC is Langevin dynamics \citep{langevin1908}. Langevin dynamics generates a Markov Chain $\{\bm{x}_n\}_{n = 1}^\infty$ using the iteration
\begin{align}
\bm{x}_{n+1} = \bm{x}_n + (\eta / 2) \cdot \nabla_{\bm{x}} \log p(\bm{x}_n) + \sqrt{\eta} \cdot \bm{\epsilon} \label{eq:langevin}
\end{align}
where $\epsilonb \sim \NN(\bm{0},\bm{I})$. $\{\bm{x}_n\}_{n = 1}^\infty$ converges to $p(\bm{x})$ in distribution for an appropriate choice of $\eta$.

To sample from a joint distribution $p(\bm{x},\bm{y})$, we may resort to Gibbs sampling \citep{geman1984}. Given a current Markov chain state $(\bm{x}_n,\bm{y}_n)$, Gibbs sampling produces $\bm{x}_{n+1}$ by sampling from $p(\bm{x} \mid \bm{y}_n)$ and $\bm{y}_{n+1}$ by sampling from $p(\bm{y} \mid \bm{x}_{n+1})$. The sampling steps may be replaced with MCMC. Hence, Gibbs sampling is useful when conditional distributions are amenable to MCMC.

\textbf{Annealed MCMC.} Despite their diversity, MCMC methods often have difficulty crossing low-density regions in high-dimensional multimodal distributions. For Langevin dynamics, at a low-density region, the score function vanishes in \eqref{eq:langevin}, resulting in a meaningless diffusion. Moreover, natural data often lies on a low-dimensional manifold. Thus, once Langevin dynamics leaves the data manifold, it becomes impossible for Langevin dynamics to find its way back.

One way to remedy this problem is to use annealing, i.e., constructing a sequence of increasingly smooth and wide distributions and running MCMC at different levels of smoothness. As smoothness is increased, disjoint modes merge, so MCMC can cross over to other modes. Annealing has been used to empower various types of MCMC \citep{geyer1995,neal2001}. In this work, we shall refer to the collection of MCMC that use annealing as annealed MCMC.

An instance of annealed MCMC is annealed Langevin dynamics (ALD) \citep{song2019score}. For a sequence of increasing noise levels $\{\sigma_i\}_{i = 1}^N$, Langevin dynamics is sequentially executed with $\int p_{\sigma_i}(\bm{x} \mid \tilde{\bm{x}}) p(\tilde{\bm{x}}) \, d\tilde{\bm{x}}$ in place of $p(\bm{x})$ in \eqref{eq:langevin} for $i = N, N - 1, \ldots, 1$. Since $p(\bm{x})$ smoothed at a large noise level has wide support and connected modes, ALD overcomes the pitfalls of vanilla Langevin dynamics. However, ALD has the drawback that thousands of iterations are required to produce a single batch of samples.

\subsection{Diffusion Models}

\textbf{Diffusion models and differential equations.} Diffusion models opened up a new avenue towards fast sampling with score functions via SDEs and ODEs \citep{song2021}. Suppose data is distributed in $\RR^d$. Given a diffusion process of data sample $\bm{x}_0 \sim p(\bm{x})$ into a sample from a simple prior noise distribution, the trajectory of data during diffusion can be described with an It\^{o} SDE
\begin{align}
d\bm{x} = \bm{f}(\bm{x},t) \, dt + g(t) \, d\bm{w} \label{eq:SDE}
\end{align}
for some drift coefficient $\bm{f} : \RR^d \times [0,T] \rightarrow \RR^d$, diffusion coefficient $g : [0,T] \rightarrow \RR$, and Brownian motion $\bm{w}$. Here, $T$ is the diffusion termination time. With initial condition $\bm{x}(0) = \bm{x}_0$, integrating \eqref{eq:SDE} from time $t = 0$ to $t = T$ produces a sample from the prior distribution.

For each diffusion SDE, there exists a corresponding reverse-SDE:
\begin{align}
d\bm{x} = [\bm{f}(\bm{x},t) - g(t)^2 \nabla_{\bm{x}} \log p_t(\bm{x})] \, dt + g(t) \, d\bar{\bm{w}} \label{eq:RSDE}
\end{align}
where $p_t(\bm{x})$ is the density of $\bm{x}(t)$ evolving according to \eqref{eq:SDE} and $\bar{\bm{w}}$ is a Brownian motion if time flows from $t = T$ to $t = 0$. Given a sample $\bm{x}_T$ from the prior distribution, integrating \eqref{eq:RSDE} with initial condition $\bm{x}(T) = \bm{x}_T$ from $t = T$ to $t = 0$ results in a sample from $p(\bm{x})$. Moreover, to each reverse-SDE, there exists a corresponding deterministic reverse-ODE
\begin{align}
d\bm{x} = \left[\bm{f}(\bm{x},t) - (1/2) \cdot g(t)^2 \nabla_{\bm{x}} \log p_t(\bm{x}) \right] \, dt
\end{align}
which also can be integrated from $t = T$ to $t = 0$ to produce samples from $p(\bm{x})$.

Diffusion models generate data by simulating the reverse of the diffusion process, i.e., by solving the reverse-S/ODE of the diffusion process. Initial works on diffusion models \citep{dickstein2015,ho2020} used computationally expensive ancestral sampling to solve the reverse differential equations. Later works discovered that using adaptive numerical integrators to solve the reverse-S/ODE could accelerate the sampling process. This led to great attention on developing better reverse-S/ODE integrators \citep{martineau2021,song2021,lu2022,karras2022,zhang2022}. Our work is orthogonal to such works as focus on finding good initialization points for integration via MCMC. Hence, a better integration technique directly translates to even better generative performance when plugged into Denoising MCMC.

\textbf{Variance exploding (VE) diffusion model.} A VE diffusion model considers the diffusion process
\begin{align}
d\bm{x} = \sqrt{\frac{d[\sigma^2(t)]}{dt}} \, d\bm{w}. \label{eq:VESDE}
\end{align}
where $\sigma(t)$ increases monotonically with $t$ from $\sigma_{\min}$ to $\sigma_{\max}$. The data distribution evolves as
\begin{align}
\textstyle p_t(\bm{x}) = \int p_{\sigma(t)}(\bm{x} \mid \tilde{\bm{x}}) p(\tilde{\bm{x}}) \, d\tilde{\bm{x}}
\end{align}
so if $\sigma_{\min}$ is sufficiently small, $p_0(\bm{x}) \approx p(\bm{x})$, and if $\sigma_{\max}$ is sufficiently large, so variance explodes, $p_T(\bm{x}) \approx \NN(\bm{x} \mid \bm{0}, \sigma_{\max}^2 \bm{I})$. If we have a score model $\s_\theta(\bm{x},\sigma)$ trained with \eqref{eq:dsm_cont}, $\nabla_{\bm{x}} \log p_t(\bm{x}) \approx \s_\theta(\bm{x}, \sigma(t))$. It follows that with $\bm{x}_T \sim \NN(\bm{x} \mid \bm{0}, \sigma_{\max}^2 \bm{I})$, we may integrate the reverse-S/ODE corresponding to \eqref{eq:VESDE} with $\bm{x}(T) = \bm{x}_T$ from $t = T$ to $t = 0$ using a score model to generate data.
In the next section, we bridge MCMC and reverse-S/ODE integrators with VE diffusion to form a novel sampling framework that improves both MCMC and diffusion models.

\section{Denoising Markov Chain Monte Carlo (DMCMC)} \label{sec:dmcmc}

From here on, we denote the data distribution as $p(\bm{x})$ and its domain as $\XX \subseteq \RR^d$. Diffusion runs from time $t = 0$ to $t = T$, and noise scale $\sigma(t)$ increases monotonically from $\sigma_{\min}$ to $\sigma_{\max}$, such that $\sigma(0) = \sigma_{\min}$ and $\sigma(T) = \sigma_{\max}$. We denote the range of $\sigma(t)$ as $\SS = [\sigma_{\min},\sigma_{\max}]$. The Gaussian perturbation kernel is denoted as $p_\sigma(\bm{x} \mid \tilde{\bm{x}}) = \NN(\bm{x} \mid \tilde{\bm{x}}, \sigma^2 \bm{I})$. The distribution of $\bm{x}(t)$ following the VE diffusion \eqref{eq:VESDE} is denoted as $p_t(\bm{x})$, and recall that $p_t(\bm{x}) = \int p_{\sigma(t)}(\bm{x} \mid \tilde{\bm{x}}) p(\tilde{\bm{x}}) \, d\tilde{\bm{x}}$.

We now develop a general framework called Denoising MCMC (DMCMC) which combines MCMC with reverse-S/ODE integrators. The construction of DMCMC is comprised of two steps. In the first step, we build MCMC on the product space $\XX \times \SS$, i.e., $\XX$ augmented by the smoothness parameter $\sigma$. Since $\sigma(t)$ is a monotone increasing function, this is equivalent to augmenting the data space with diffusion time $t$. In the second step, we incorporate denoising steps, where we denoise MCMC samples via reverse-S/ODE integrators.

\subsection{Construction Step 1: MCMC on the Product Space $\mathcal{X} \times \mathcal{S}$}

Suppose $p(\bm{x})$ is a high-dimensional multimodal distribution, supported on a low-dimensional manifold. If the modes are separated by wide low-density regions, MCMC can have difficulty moving between the modes. Indeed, convergence time for such distributions can grow exponential in dimension $d$ \citep{roberts2001}. Intuitively, for MCMC to move between disjoint modes, the Markov Chain would have to step off the data manifold. However, once MCMC leaves the data manifold, the density or the score vanishes. Then, most random directions produced by the proposal distribution do not point to the manifold. Thus, MCMC gets lost in the ambient space, whose volume grows exponentially in $d$.

Annealing via Gaussian smoothing, used in both ALD and VE diffusion, circumvents this problem. As $p(\bm{x})$ smoothed with perturbation kernel $p_\sigma(\bm{x} \mid \tilde{\bm{x}})$ of increasing $\sigma$, the modes of $p(\bm{x})$ grow wider and start to connect. Thus, MCMC can easily transition between modes. However, running MCMC in the manner of ALD is inefficient since we do not know how many iterations within each noise level is sufficient. To address this problem, we propose to augment $\XX$ with the smoothness scale $\sigma$ and run MCMC in the product space $\XX \times \SS$ such that MCMC automatically controls the value of $\sigma$. Below, we formally describe MCMC on $\XX \times \SS$.

Let us define the $\sigma$-conditional distribution
\begin{align}
\textstyle \hat{p}(\bm{x} \mid \sigma) \coloneqq \int p_\sigma(\bm{x} \mid \tilde{\bm{x}}) p(\tilde{\bm{x}}) \, d\bm{\tilde{x}}. \label{eq:cond}
\end{align}
We also define a prior $\hat{p}(\sigma)$ on $\SS$. Then by the Bayes' Rule,
\begin{align}
\hat{p}(\bm{x},\sigma) = \hat{p}(\bm{x} \mid \sigma) \cdot \hat{p}(\sigma).
\end{align}
Here, $\hat{p}(\sigma)$ reflects our preference for how much time we want the MCMC chain to stay at a particular level of $\sigma$. MCMC with $\hat{p}(\bm{x},\sigma)$ will produce samples $\{(\bm{x}_n,\sigma_n)\}$ in $\XX \times \SS$ such that
\begin{align}
\sigma_n \sim \hat{p}(\sigma), \qquad \bm{x}_n \sim \hat{p}(\bm{x} \mid \sigma_n). \label{eq:sample_dist}
\end{align}
Hence, if $\sigma_n \gg \sigma_{\min}$, $\bm{x}_n$ will be a noisy sample, i.e., a sample corrupted with Gaussian noise of variance $\sigma_n^2$, and if $\sigma_n \approx \sigma_{\min}$, $\bm{x}_n$ will resemble a sample from $p(\bm{x})$.

Since our goal is to generate samples from $p(\bm{x})$, na\"{i}vely, we can keep samples $(\bm{x}_n,\sigma_n)$ with $\sigma_n \approx \sigma_{\min}$ and discard other samples. However, this could lead to a large waste of computation resources. In the next section, we incorporate reverse-S/ODE integrators to avert this problem.

\subsection{Construction Step 2: Incorporating Denoising Steps}

Let us recall that integrating the reverse-S/ODE for the VE diffusion SDE Eq. (\ref{eq:VESDE}) from time $t = T$ to $t = 0$ sends samples from $p_T(\bm{x})$ to samples from $p_0(\bm{x}) \approx p(\bm{x})$. In general, integrating the reverse-SDE or ODE from time $t = t_2$ to $t = t_1$ for $t_1 < t_2$ sends samples from $p_{t_2}(\bm{x})$ to samples from $p_{t_1}(\bm{x})$ \citep{song2021}. We use this fact to denoise MCMC samples from $\hat{p}(\bm{x},\sigma)$.

Suppose we are given a sample $(\bm{x}_n,\sigma_n) \sim \hat{p}(\bm{x},\sigma)$. With $t_n \coloneqq \sigma^{-1}(\sigma_n)$, \eqref{eq:sample_dist} tells us
\begin{align}
\bm{x}_n \sim p_{t_n}(\bm{x})
\end{align}
so integrating the reverse-S/ODE with initial condition $\bm{x}(t_n) = \bm{x}_n$ from $t = t_n$ to $t = 0$ produces a sample from $p_0(\bm{x}) \approx p(\bm{x})$. Here, we note that any reverse-S/ODE solver may be used to carry out the integration.

Given an MCMC chain $\{(\bm{x}_n,\sigma_n)\}$ in $\XX \times \SS$, MCMC is biased towards high-density regions of $\XX$, so the sequence $\{\bm{x}_n\}$ will generally stay close to the data manifold, except when traversing between disjoint modes. This means $\sigma_n \ll \sigma_{\max}$ for most $n$, or in other words, $t_n \ll T$ for most $n$. So, the average length of integration intervals will tend to be much shorter than $T$. Thus, a numerical integrator can integrate the reverse-S/ODE with less truncation error given the same computation budget \citep{numanalysis}. Equivalently, less computation budget is required to reach the same truncation error level. This idea is illustrated in Figure \ref{fig:dmcmc_example}.

\section{Denoising Langevin Gibbs (DLG)}

In Section \ref{sec:dmcmc}, we described an abstract framework, DMCMC, for accelerating score-based sampling by combining MCMC and reverse-S/ODE integrators. We now develop a concrete instance of DMCMC. As the second construction step of DMCMC is simple, we only describe the first step.

Na\"{i}vely, we could extend denoising score matching \eqref{eq:dsm_orig} to estimate the score $\hat{s}_\theta(\bm{x},\sigma) : \RR^d\times\RR \rightarrow \RR^d\times\RR$ of $\hat{p}(\bm{x},\sigma)$ and apply Langevin dynamics in the first step of DMCMC. But, this would prevent us from using pre-trained score models, as we would have to solve (for some small $\nu > 0$)
\begin{align}
\min_\theta \EE_{\bm{\epsilon} \sim \NN(\bm{0}_{d+1},\nu^2 \bm{I}_{d+1})} \EE_{\hat{p}(\bm{x}, \sigma)} [\|\hat{s}_\theta(\bm{x}-\epsilon_{1:d},\sigma-\epsilon_{d+1}) - \bm{\epsilon}/\nu^2\|_2^2]
\end{align}
which differs from \eqref{eq:dsm_cont}. Gibbs sampling provides a simple path around this problem. Let us recall that given a previous MCMC iterate $(\bm{x}_n,\sigma_n)$, Gibbs sampling proceeds by alternating between an $\bm{x}$ update step $\bm{x}_{n+1} \sim \hat{p}(\bm{x} \mid \sigma_n)$ and a $\sigma$ update step $\sigma_{n+1} \sim \hat{p}(\sigma \mid \bm{x}_{n+1})$. Below, we describe our score-based sampling algorithm, Denoising Langevin Gibbs (DLG).

\textbf{Updating $\bm{x}$.} Suppose we are given an MCMC iterate $(\bm{x}_n,\sigma_n)$ and a score model $s_\theta(\bm{x},\sigma)$ from \eqref{eq:dsm_cont}. We generate $\bm{x}_{n+1}$ by a Langevin dynamics step on $\hat{p}(\bm{x} \mid \sigma_n)$. Specifically, by \eqref{eq:cond},
\begin{align}
\nabla_{\bm{x}} \log \hat{p}(\bm{x} \mid \sigma_n) \approx s_\theta(\bm{x}, \sigma_n) \label{eq:dlg_lang}
\end{align}
and so an Langevin dynamics update on $\bm{x}$, according to \eqref{eq:langevin} is
\begin{align}
\bm{x}_{n+1} = \bm{x}_n + (\eta/2) \cdot s_\theta(\bm{x}_n,\sigma_n) + \sqrt{\eta} \cdot \bm{\epsilon}
\end{align}
for $\bm{\epsilon} \sim \NN(\bm{0},\bm{I})$. Here, we call $\eta$ the step size.

\textbf{Updating $\sigma$.} We now have $\bm{x}_{n+1}$ and need to sample $\sigma_{n+1} \sim \hat{p}(\sigma \mid \bm{x}_{n+1})$. To this end, we first train a DNN noise level classifier $q_{\phi}(\sigma \mid \bm{x})$ to approximate $\hat{p}(\sigma \mid \bm{x})$ by solving
\begin{align}
\max_{\phi} \EE_{\hat{p}(\bm{x},\sigma)} [\log q_{\phi}(\sigma \mid \bm{x})].
\end{align}
Specifically, we discretize $[\sigma_{\min},\sigma_{\max}]$ into $M$ levels $\tau_1 = \sigma_{\min} < \tau_2 < \cdots < \tau_M = \sigma_{\max}$. Given $\tau_m$ where $1 \leq m \leq M$, $m$ serves as the label and clean training data corrupted by Gaussian noise of variance $\tau_m^2$ serves as the classifier input. The classifier is trained to predict $m$ by minimizing the cross entropy loss. Having trained a noise level classifier, we sample $\sigma_{n+1}$ by drawing an index $m$ according to the classifier output probability for $\bm{x}_{n+1}$ and setting $\sigma_{n+1} = \tau_m$. In practice, using the index of largest probability worked fine. We denote this process as $\sigma_{n+1} \sim q_{\phi}(\sigma \mid \bm{x}_{n+1})$.

\subsection{Practical Considerations}

\textbf{Computation cost of $\sigma$ prediction.} We found that using shallow neural networks for the noise classifier $q_{\phi}$ was sufficient to accelerate sampling. Concretely, using a neural net with four convolution layers and one fully connected layer as the classifier, one evaluation of $q_{\phi}$ was around $100 \sim 1000$ times faster than one evaluation of the score model $\s_\theta$. So, when comparing sampling methods, we only count the number of score function evaluations (NFE). We also note that the training time $q_{\phi}$ was negligible compared to the training time of $s_\theta$. For instance, on CelebA-HQ-256, training $q_{\phi}$ with the aforementioned architecture for 100 epochs took around 15 minutes on an RTX 2080 Ti.

\textbf{Initialization points for DLG.} Theoretically, MCMC chain $\{(\bm{x}_n,\sigma_n)\}_{n = 1}^\infty$ will converge to $\hat{p}(\bm{x},\sigma)$ regardless of the initialization point $(\bm{x}_0,\sigma_0)$. However, we found it was beneficial to set $\bm{x}_0$ close to the image manifold and set $\sigma_0 \sim q_{\phi}(\sigma \mid \bm{x}_0)$. Indeed, theory also shows that setting initialization points close to the stationary distribution could significantly accelerate convergence of the Markov chain \citep{dalalyan2017,dwivedi2019}. In practice, we set $\bm{x}_0$ by generating a sample starting from a noise distribution, adding Gaussian noise of variance $0.25$, and running Gibbs sampling for a few iterations. The NFE involved in generating $\bm{x}_0$ is included in the final per-sample average NFE computation for DLG when comparing methods in Section \ref{sec:exp}. But, we note that this cost vanishes in the limit of infinite sample size.

\textbf{Reducing autocorrelation.} Autocorrelation in MCMC chains, i.e., correlation between consecutive samples in the MCMC chain, could reduce the sample diversity of MCMC. A typical technique to reduce autocorrelation is to use every $n_{skip}$-th samples of the MCMC chain for some $n_{skip} > 1$. For DMCMC, this means we denoise every $n_{skip}$-th sample. So, if we use $n_{den}$ NFE to denoise MCMC samples, the average NFE for generating a single sample is around $n_{skip} + n_{den}$.

\textbf{Choosing iterates to apply denoising.} The MCMC chain can be partitioned into blocks which consist of $n_{skip}$ consecutive samples. Using every $n_{skip}$-th sample of the MCMC chain corresponds to denoising the last iterate of each block. Instead, to further shorten the length of integration, within each block, we apply denoising to the sample of minimum noise scale $\sigma$.

\textbf{Choice of prior $\hat{p}(\sigma)$.} We use $\hat{p}(\sigma) \propto 1/\sigma$ to drive the MCMC chain towards small values of $\sigma$.

\begin{figure}[t]
\centering
\includegraphics[width=1.0\linewidth]{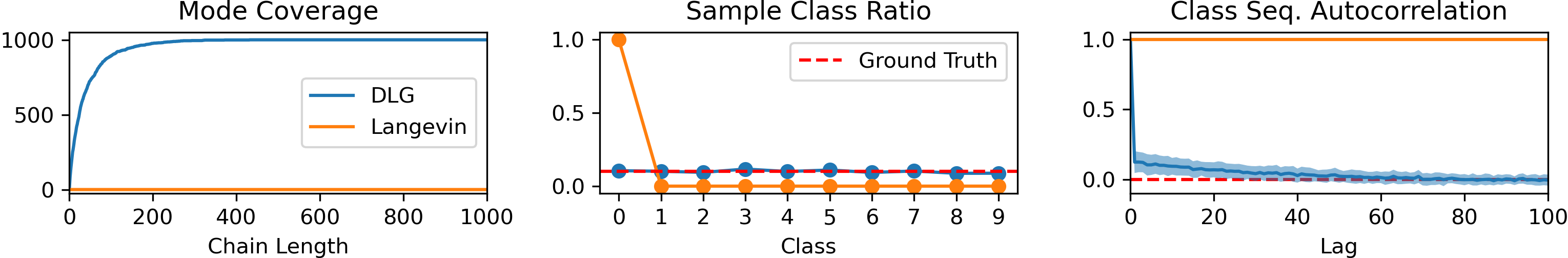} \\
\caption{Mixing analysis of DLG with a mixture of Gaussians.}
\label{fig:mixing}
\end{figure}

\begin{figure}[t]
\centering
\includegraphics[width=1.0\linewidth]{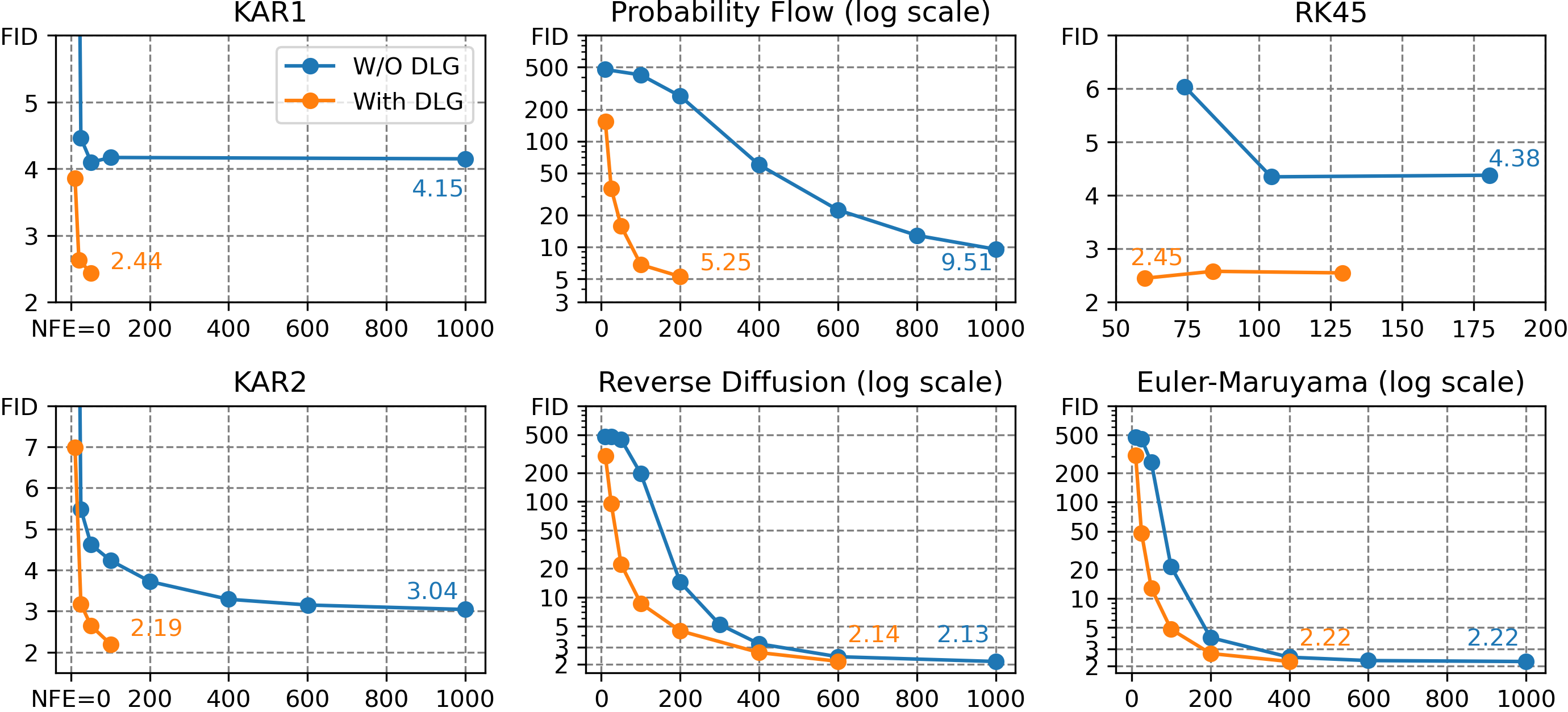}
\caption{Sampling acceleration of DLG on CIFAR10. FID of notable points are written in the corresponding color. \textbf{Top row}: deterministic integrators. \textbf{Bottom row}: stochastic integrators.}
\label{fig:cifar10_acc}
\end{figure}

\section{Experiments} \label{sec:exp}

\subsection{Mixing of DMCMC Chains}

For DMCMC to successfully generate diverse high-quality data, DMCMC must mix, i.e., traverse between disjoint modes of $p(\bm{x})$. We provide experimental evidence that DMCMC indeed mixes as a consequence of running MCMC in the product space $\XX \times \SS$. Specifically, we run fifty Langevin dynamics chains and fifty DLG chains on a mixture of Gaussians (MoG) with $1k$ modes at CIFAR10 images. All chains are initialized at a single mode. For each method, we compute the mode coverage of the samples, the class distribution of the samples, and the autocorrelation of sample image class sequence. Since the noise conditional score function can be calculated analytically for MoGs, this setting decouples sampler performance from score model approximation error.

Figure \ref{fig:mixing} shows the results. In the left panel, we observe that Langevin dynamics is unable to escape the initial mode. Increasing the step size $\eta$ of Langevin dynamics caused the chain to diverge. On the other hand, DLG successfully captures all modes of the distribution. DLG samples cover all $1k$ modes at chain length $432$. Middle panel provides evidence that DLG samples correctly reflect the statistics of the data distribution. Finally, the right panel indicates that the DLG chain moves freely between classes, i.e., distant modes. These observations validate our claim that DLG mixes well.

\subsection{Accelerating Image Generation with Score Networks}

We compare six integrators with and without DLG on CIFAR10 and CelebA-HQ-256 image generation. The deterministic integrators are: the deterministic integrator of \citet{karras2022} (KAR1), the probability flow integrator of \citet{song2021}, and the RK45 solver. The stochastic integrators are: the stochastic integrator of \citet{karras2022} (KAR2), the reverse diffusion integrator of \citet{song2021}, and the Euler-Maruyama method. We use the Fr\'{e}chet Inception Distance (FID) \citep{heusel2017} to measure sample quality. For CIFAR10, we generate $50k$ samples, and for CelebA-HQ-256, we generate $10k$ samples. We use pre-trained score models of \citet{song2021}.

\begin{figure}[t]
\centering
\includegraphics[width=1.0\linewidth]{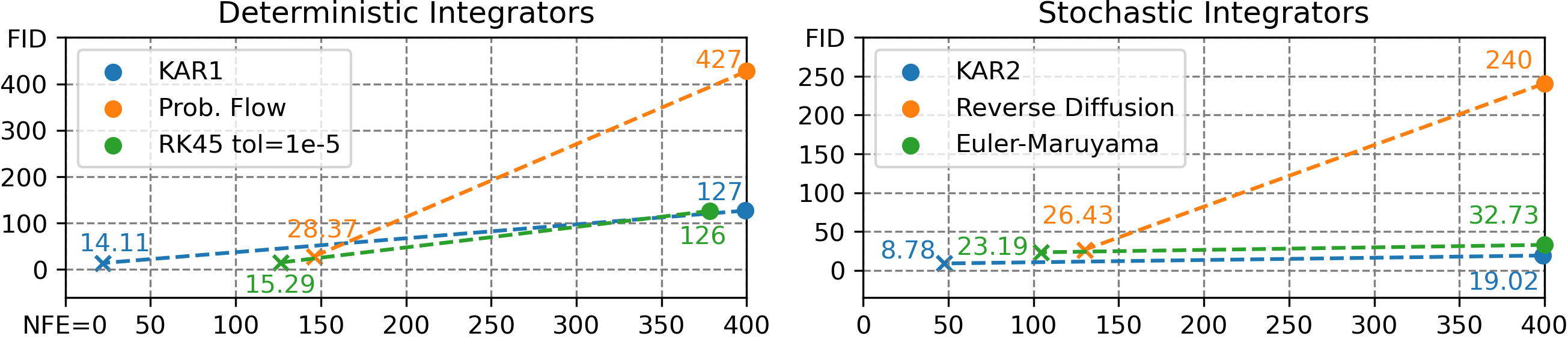}
\caption{Sampling acceleration of DLG on CelebA-HQ-256. A dot indicates an integrator without DLG, and a cross of the same color indicates corresponding integrator combined with DLG. Dotted lines indicate performance improvement due to DLG.}
\label{fig:celeba_acc}
\end{figure}

\begin{figure}[t]
\centering
\includegraphics[width=0.497\linewidth]{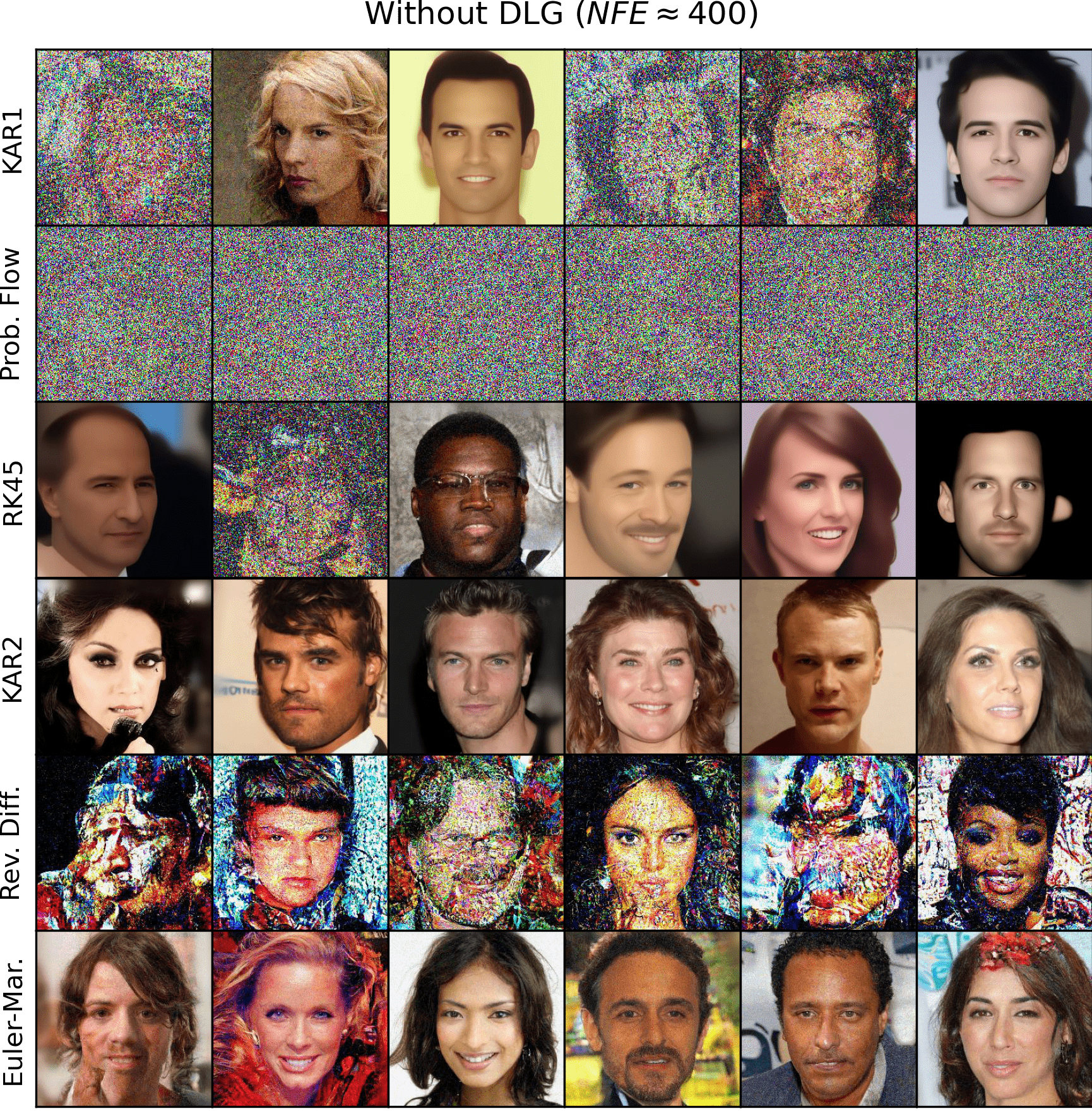}
\hspace{0.5mm}
\includegraphics[width=0.485\linewidth]{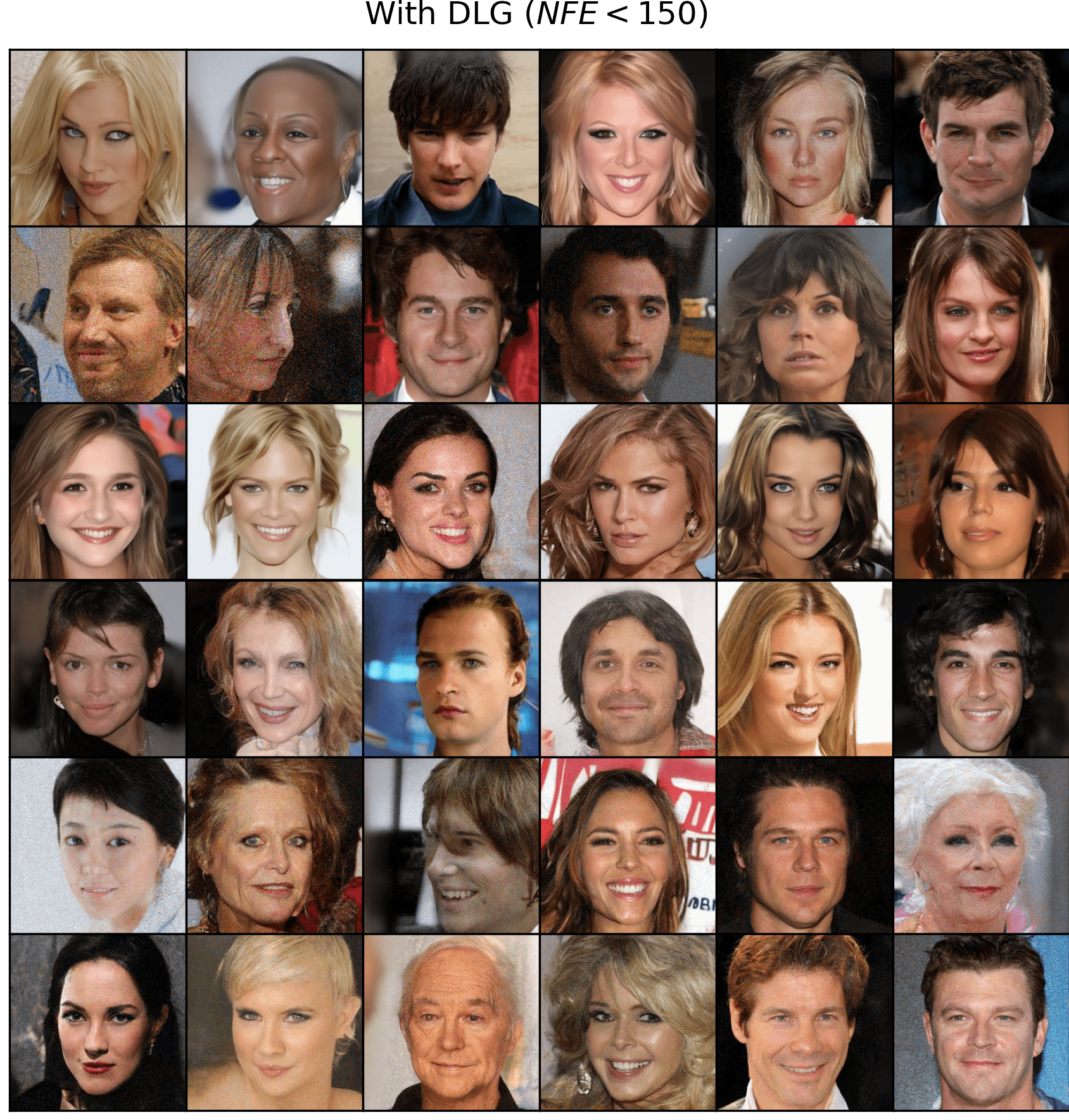}
\caption{Non-cherry-picked samples on CelebA-HQ-256 using the settings for Fig. \ref{fig:celeba_acc}. Each row shows samples for an integrator without (left col.) and with (right col.) DLG.}
\label{fig:samples}
\end{figure}
 
\textbf{CIFAR10.} Figure \ref{fig:cifar10_acc} shows the results on CIFAR10. We make two important observations. First, DLG successfully accelerates all six integrators by a non-trivial margin. In particular, if an integrator without DLG already performs well, the integrator combined with DLG outperforms other integrators combined with DLG. For instance, compare the results for KAR1 with those of other deterministic integrators. Second, DLG improves the performance lower bound for some deterministic integrators. Specifically, while the performance of KAR1 and RK45 saturates at around $4$ FID, KAR1 and RK45 combined with DLG achieve better results around $2.4$ FID.

\textbf{CelebA-HQ-256.} Figure \ref{fig:celeba_acc} shows the results on CelebA-HQ-256. We observe that DLG improves computational efficiency and sample quality simultaneously. Indeed, in Figure \ref{fig:samples}, we observe remarkable improvements in sample quality despite using fewer NFE. This demonstrates the scalability of DLG to generating high-resolution images. We also note that we did not perform an exhaustive search of DLG hyper-parameters for CelebA-HQ-256, so fine-tuning could yield better results.

\textbf{Achieving SOTA.} DLG combined with KAR1 sets a new SOTA record for CIFAR10 in the limited number of NFE setting: $3.86$ FID with $10.11$ NFE and $2.63$ FID with $20.11$ NFE which beats the results of \citet{zhang2022}, $4.17$ FID with $10$ NFE and $2.86$ FID with $20$ NFE. DLG combined with KAR2 sets a new record on CelebA-HQ-256 among score-based models: $6.99$ FID with $158.96$ NFE which beats the current best result of \citet{kim2022}, $7.16$ FID with $4000$ NFE.

\begin{figure}[t]
\centering
\includegraphics[width=1.0\linewidth]{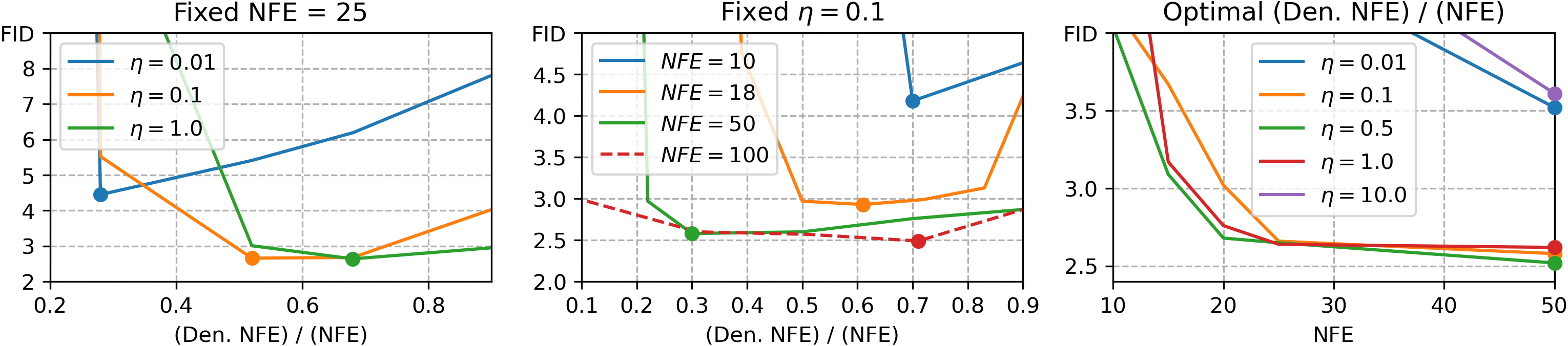}
\caption{Ablation study of DLG with the deterministic integrator of \citet{karras2022}. Dots indicate the points of lowest FID.}
\label{fig:ablation}
\end{figure}

\subsection{Ablation Study}

In this section, we perform an ablation study of the components of DLG. Based on the observations, we construct a generic strategy for choosing the hyper-parameters. Given a reverse-S/ODE integrator, DLG is determined by three hyper-parameters: total NFE per sample $n$, NFE spent on denoising samples $n_{den}$, and Langevin dynamics step size $\eta$. We fix the integrator to be the deterministic sampler of \citet{karras2022} and observe the effect of each hyper-parameter on CIFAR10 image generation. We observed the same general trend with other samplers.

\textbf{$\bm{\eta}$ vs. $\bm{n}_{\bm{den}}/ \bm{n}$.} In the left panel of Figure \ref{fig:ablation}, we fix NFE and vary $\eta$ and $n_{den}/n$. $n_{den}$ governs individual sample quality, and $n_{skip} = n - n_{den}$ governs sample diversity. Thus, we observe optimal FID is achieved when $n_{den}/n$ has intermediate values, not extreme values near $0$ or $1$. Also, lower $n_{den}/n$ is needed to attain optimality for lower $\eta$. This is because lower $\eta$ means the MCMC chain travels closer to the image manifold at the cost of slower mixing.

\textbf{NFE vs. $\bm{n}_{\bm{den}}/ \bm{n}$.} In the middle panel of Figure \ref{fig:ablation}, we fix $\eta$ and vary NFE and $n_{den}/n$. We observe two trends. First, in the small NFE regime, where $10 \leq \text{NFE} \leq 50$, it is beneficial to decrease $n_{den}/n$ as NFE increases. Second, in the large NFE regime, where $\text{NFE} > 50$, it is beneficial to increase $n_{den}/n$ as NFE increases. This is because if $n_{skip}$ is sufficiently large, MCMC chain starts producing essentially independent samples, so increasing $n_{skip}$ further provides no gain.

Also, as we increase NFE, the set of $n_{den}/n$ which provides near-optimal performance becomes larger. Moreover, we see that most of the time, optimal FID is achieved when $n_{den} / n > 0.5$, i.e., when $n_{den} > n_{skip}$. So, a reasonable strategy for choosing $n_{den}$ given $\eta$ and NFE budget $n$ is to find smallest $n_{skip}$ which produces visually distinct samples, and then allocate $n_{den} = n - n_{skip}$.

\textbf{$\bm{\eta}$ vs. NFE.} In the right panel of Figure \ref{fig:ablation}, we choose optimal (in terms of FID) $n_{den}/n$ for each combination of $\eta$ and NFE. We see choosing overly small or large $\eta$ leads to performance degradation. If $\eta$ is within a certain range, we obtain similarly good performance. In the case of CIFAR10, we found it reasonable to set $\eta \in [0.05,1.0]$.

To choose $\eta$ for data of general dimension $d$, we define a value $\kappa \coloneqq \eta / \sqrt{d}$ called displacement per dimension. If we see \eqref{eq:dlg_lang}, Gaussian noise of zero mean and variance $\eta$ is added to the sample at each update step. Gaussian annulus theorem tells us that a high-dimensional Gaussian noise with zero mean and variance $\eta$ has Euclidean norm approximately $\eta \sqrt{d}$. So, the average displacement of the sample per dimension by the random noise is around $\eta / \sqrt{d}$. Since $\kappa$ is a dimension-independent value, given $\kappa$ and $d$, we can set $\eta = \sqrt{d} \kappa$. On CIFAR10, we have $\eta \in [0.05,1.0]$, which translates to $\kappa \in [0.0009, 0.018]$. This means, on CelebA-HQ-256, we can choose $\eta \in [0.4,8.0]$. If the sampler was inefficient, we chose a smaller $\kappa$ to trade-off diversity for sample quality.

\section{Conclusion}

In this work, we proposed DMCMC which combines MCMC with reverse-S/ODE integrators. This has led to significant improvements for both MCMC and diffusion models. For MCMC, DMCMC allows Markov chains to traverse between disjoint modes. For diffusion models , DMCMC accelerates sampling by reducing the average integration interval length of reverse-S/ODE. We developed a practical instance of DMCMC, called DLG, which uses Langevin dynamics along with Gibbs sampling as the choice of MCMC. We demonstrated the practicality and scalability of DLG through various experiments. In particular, DLG achieved state-of-the-art results on CIFAR10 and CelebA-HQ-256. Overall, our work opens up an orthogonal approach to accelerating score-based sampling. We leave exploration of other kinds of MCMC or diffusion process in DMCMC as future work.

\bibliographystyle{iclr2023_conference}

\begin{thebibliography}{30}
\providecommand{\natexlab}[1]{#1}
\providecommand{\url}[1]{\texttt{#1}}
\expandafter\ifx\csname urlstyle\endcsname\relax
  \providecommand{\doi}[1]{doi: #1}\else
  \providecommand{\doi}{doi: \begingroup \urlstyle{rm}\Url}\fi

\bibitem[Dalalyan(2017)]{dalalyan2017}
Arnak~S. Dalalyan.
\newblock Theoretical guarantees for approximate sampling from smooth and
  log-concave densities.
\newblock \emph{Journal of the Royal Statistical Society}, 79\penalty0
  (3):\penalty0 651--676, 2017.

\bibitem[Dhariwal \& Nichol(2021)Dhariwal and Nichol]{dhariwal2021}
Prafulla Dhariwal and Alexander~Quinn Nichol.
\newblock Diffusion models beat {GAN}s on image synthesis.
\newblock In \emph{NeurIPS}, 2021.

\bibitem[Dwivedi et~al.(2019)Dwivedi, Chen, J.Wainwright, and Yu]{dwivedi2019}
Raaz Dwivedi, Yuansi Chen, Martin J.Wainwright, and Bin Yu.
\newblock Log-concave sampling: {Metropolis-Hastings} algorithms are fast.
\newblock \emph{Journal of Machine Learning Research}, 20:\penalty0 1--42,
  2019.

\bibitem[Geman \& Geman(1984)Geman and Geman]{geman1984}
Stuart Geman and Donald Geman.
\newblock Stochastic relaxation, {Gibbs} distributions, and the {Bayesian}
  restoration of images.
\newblock \emph{IEEE Transactions on Pattern Analysis and Machine
  Intelligence}, 6\penalty0 (6):\penalty0 721--741, 1984.

\bibitem[Geyer \& Thompson(1995)Geyer and Thompson]{geyer1995}
Charles~J. Geyer and Elizabeth~A. Thompson.
\newblock Annealing {Markov} chain {Monte} {Carlo} with applications to
  ancestral inferences.
\newblock \emph{Journal of the American Statistical Association}, 90\penalty0
  (431):\penalty0 909--920, 1995.

\bibitem[Hastings(1970)]{hastings1970}
Wilfred~K. Hastings.
\newblock Monte {Carlo} sampling methods using {Markov} chains and their
  application.
\newblock \emph{Biometrika}, 57:\penalty0 97--109, 1970.

\bibitem[Heusel et~al.(2017)Heusel, Ramsauer, Unterthiner, Nessler, and
  Hochreiter]{heusel2017}
Martin Heusel, Hubert Ramsauer, Thomas Unterthiner, Bernhard Nessler, and Sepp
  Hochreiter.
\newblock {GANs} trained by a two time-scale update rule converge to a local
  {Nash} equilibrium.
\newblock In \emph{NIPS}, 2017.

\bibitem[Ho et~al.(2020)Ho, Jain, and Abbeel]{ho2020}
Jonathan Ho, Ajay Jain, and Pieter Abbeel.
\newblock Denoising diffusion probabilistic models.
\newblock In \emph{NeurIPS}, 2020.

\bibitem[Jolicoeur-Martineau et~al.(2021)Jolicoeur-Martineau, Li,
  Pich\'{e}-Taillefer, Kachman, and Mitliagkas]{martineau2021}
Alexia Jolicoeur-Martineau, Ke~Li, R\'{e}mi Pich\'{e}-Taillefer, Tal Kachman,
  and Ioannis Mitliagkas.
\newblock Gotta go fast when generating data with score-based models.
\newblock \emph{arxiv preprint arXiv:2105.14080}, 2021.

\bibitem[Karras et~al.(2018)Karras, Aila, Laine, and Lehtinen]{karras2018}
Tero Karras, Timo Aila, Samuli Laine, and Jaakko Lehtinen.
\newblock Progressive growing of {GANs} for improved quality, stability, and
  variation.
\newblock In \emph{ICLR}, 2018.

\bibitem[Karras et~al.(2022)Karras, Aittala, Aila, and Laine]{karras2022}
Tero Karras, Miika Aittala, Timo Aila, and Samuli Laine.
\newblock Elucidating the design space of diffusion-based generative models.
\newblock \emph{arxiv preprint arXiv:2206.00364}, 2022.

\bibitem[Kim et~al.(2022)Kim, Shin, Song, Kang, and Moon]{kim2022}
Dongjun Kim, Seungjae Shin, Kyungwoo Song, Wanmo Kang, and Il-Chul Moon.
\newblock Soft truncation: A universal truncation technique of score-based
  diffusion for high precision score estimation.
\newblock In \emph{ICML}, 2022.

\bibitem[Kingma \& Ba(2015)Kingma and Ba]{kingma2015}
Diedrik~P. Kingma and Jimmy Ba.
\newblock Adam: A method for stochastic optimization.
\newblock 2015.

\bibitem[Krizhevsky(2009)]{cifar}
Alex Krizhevsky.
\newblock Learning multiple layers of features from tiny images.
\newblock Technical report, University of Toronto, 2009.

\bibitem[Langevin(1908)]{langevin1908}
Paul Langevin.
\newblock On the theory of {Brownian} motion.
\newblock 1908.

\bibitem[Levy et~al.(2018)Levy, Hoffman, and Sohl-Dickstein]{levy2018}
Daniel Levy, Matthew~D. Hoffman, and Jascha Sohl-Dickstein.
\newblock Generalizing {Hamiltonian} {Monte} {Carlo} with neural networks.
\newblock In \emph{ICLR}, 2018.

\bibitem[Lu et~al.(2022)Lu, Zhou, Bao, Chen, Li, and Zhu]{lu2022}
Cheng Lu, Yuhao Zhou, Fan Bao, Jianfei Chen, Chongxuan Li, and Jun Zhu.
\newblock Dpm-solver: A fast {ODE} solver for diffusion probabilistic sampling
  in around 10 steps.
\newblock \emph{arxiv preprint arXiv:2206.00927}, 2022.

\bibitem[Metropolis et~al.(1953)Metropolis, Rosenbluth, Rosenbluth, Teller, and
  Teller]{metropolis1953}
Nicholas Metropolis, Arianna~W. Rosenbluth, Marshall~N. Rosenbluth, Augusta~H.
  Teller, and Edward Teller.
\newblock Equations of state calculations by fast computing machines.
\newblock \emph{J. Chem. Phys.}, 21\penalty0 (6):\penalty0 1087--1092, 1953.

\bibitem[Neal(2001)]{neal2001}
Radford~M. Neal.
\newblock Annealed importance sampling.
\newblock \emph{Statistics and Computing}, 11:\penalty0 125--139, 2001.

\bibitem[Neal(2011)]{neal2011}
Radford~M. Neal.
\newblock {MCMC} using {Hamiltonian} dynamics.
\newblock \emph{Handbook of {Markov} Chain {Monte} {Carlo}}, 2\penalty0
  (11):\penalty0 2, 2011.

\bibitem[Roberts \& Rosenthal(2001)Roberts and Rosenthal]{roberts2001}
Gareth~O. Roberts and Jeffrey~S. Rosenthal.
\newblock Optimal scaling for various {Metropolis-Hastings} algorithms.
\newblock \emph{Statistical Science}, 16\penalty0 (4):\penalty0 351--367, 2001.

\bibitem[Sohl-Dickstein et~al.(2015)Sohl-Dickstein, Weiss, Maheswaranathan, and
  Ganguli]{dickstein2015}
Jascha Sohl-Dickstein, Eric~A. Weiss, Niru Maheswaranathan, and Surya Ganguli.
\newblock Deep unsupervised learing using nonequilibrium thermodynamics.
\newblock In \emph{ICML}, 2015.

\bibitem[Song et~al.(2021{\natexlab{a}})Song, Meng, and Ermon]{song2021ddim}
Jiaming Song, Chenlin Meng, and Stefano Ermon.
\newblock Denoising diffusion implicit models.
\newblock In \emph{ICLR}, 2021{\natexlab{a}}.

\bibitem[Song \& Ermon(2019)Song and Ermon]{song2019score}
Yang Song and Stefano Ermon.
\newblock Generative modeling by estimating gradients of the data distribution.
\newblock In \emph{NeurIPS}, 2019.

\bibitem[Song \& Kingma(2021)Song and Kingma]{song2021ebm}
Yang Song and Diederik~P. Kingma.
\newblock How to train your energy-based models.
\newblock \emph{arxiv preprint arXiv:2101.03288}, 2021.

\bibitem[Song et~al.(2021{\natexlab{b}})Song, Sohl-Dickstein, Kingma, Kumar,
  Ermon, and Poole]{song2021}
Yang Song, Jascha Sohl-Dickstein, Diederik~P. Kingma, Abhishek Kumar, Stefano
  Ermon, and Ben Poole.
\newblock Score-based generative modeling through stochastic differential
  equations.
\newblock In \emph{ICLR}, 2021{\natexlab{b}}.

\bibitem[Stoer \& Bulisch(2002)Stoer and Bulisch]{numanalysis}
Josef Stoer and Roland Bulisch.
\newblock \emph{Introduction to Numerical Analysis}, volume~12.
\newblock Springer Science+Business Media New York, 2002.

\bibitem[Vincent(2011)]{vincent2011}
Pascal Vincent.
\newblock A connection between score matching and denoising autoencoders.
\newblock \emph{Neural Computation}, 23\penalty0 (7):\penalty0 1661--1674,
  2011.

\bibitem[Welling \& Teh(2011)Welling and Teh]{welling2011}
Max Welling and Yee~Whye Teh.
\newblock Bayesian learning via stochastic gradient {Langevin} dynamics.
\newblock 2011.

\bibitem[Zhang \& Chen(2022)Zhang and Chen]{zhang2022}
Qinsheng Zhang and Yongxin Chen.
\newblock Fast sampling of diffusion models with exponential integrator.
\newblock \emph{arxiv preprint arXiv:2204.13902}, 2022.

\end{thebibliography}


\newpage

\newpage

\appendix

\section{Detailed Experiment Settings} \label{append:settings}

\textbf{Device.} We use an RTX 2080 Ti or two Quadro RTX 6000 depending on the required VRAM.

\textbf{Codes.} For probability flow, RK45, reverse diffusion, and Euler-Maruyama integrators, we modify the code provided by \citet{song2021} in the GitHub repository \url{https://github.com/yang-song/score_sde_pytorch}. For KAR1 and KAR2, since \citet{karras2022} did not release their implementation of the samplers, we used our implementation based on their paper. For evaluation, we use the FID implementation provided in the GitHub repository \url{https://github.com/mseitzer/pytorch-fid}.

\textbf{Datasets.} We use the CIFAR10 dataset \citep{cifar} and the CelebA-HQ-256 dataset \citep{karras2018}.

\textbf{Data processing.} All data are normalized into the range $[0,1]$. Following \citet{song2021}, for all methods, a denoising step using Tweedie's denoising formula is applied at the end of the sampling process.

\textbf{Noise predictor network $q_\phi$.} The noise predictor network $q_\phi$ has four convolution layers followed by a fully connected layer. The convolution layers have channels 32, 64, 128, 256, and $(\sigma_{\min},\sigma_{\max})$ is discretized into $1k$ points $\sigma_{\min} (\sigma_{\max}/\sigma_{\min})^t$ for $t$ spaced evenly on $[0,1]$. On CIFAR10, $q_\phi$ is trained for 200 epochs, and on CelebA-HQ-256, $q_\phi$ is trained for 100 epochs. We use the Adam optimizer \citep{kingma2015} with learning rate $0.001$.

\textbf{Mixture of Gaussians.} MoG has $1k$ modes at randomly sampled CIFAR10 training set images. For Langevin dynamics, we use step size $\eta = 0.0001$. Using a larger step size caused the Langevin dynamics chain to diverge. For DLG, we use step size $\eta = 1.0$, $n_{skip} = 1$, $n_{den} = 20$, and the reverse diffusion integrator. For each method, all chains were initialized at a single mode to test mixing capabilities in the worst-case scenario.

\textbf{CIFAR10 image generation.} We use a pre-trained NCSN++ (cont.) score model provided by \citet{song2021}. For the baseline methods, we use the recommended settings. For DLG, the chain was initialized by generating samples with the deterministic integrator of \citet{karras2022} using 37 NFE, adding Gaussian noise of variance $0.25$, and running 20 iterations of Langevin-Gibbs. Table \ref{tab:cifar10_param} lists the hyper-parameters for DLG and the corresponding FID used to produce Figure \ref{fig:cifar10_acc}.

\begin{table*}[h!]
\centering
\resizebox{0.8\textwidth}{!}{
\begin{tabular}{c c c c c c c c c c c c}
\toprule
\multicolumn{4}{c}{\textbf{KAR1}} & \multicolumn{4}{c}{\textbf{Probability Flow}} & \multicolumn{4}{c}{\textbf{RK45}} \\
\cmidrule(lr){1-4} \cmidrule(lr){5-8} \cmidrule(lr){9-12}
$n_{den}$ & $n_{skip}$ & $\eta$ & FID & $n_{den}$ & $n_{skip}$ & $\eta$ & FID & $n_{den}$ & $n_{skip}$ & $\eta$ & FID \\
\cmidrule{1-12}
9  & 1  & $0.5$ & $3.86$ & 9   & 1  & $0.01$ & $153.48$ & $59.98$ & 10 & $0.5$ & $2.45$ \\
17 & 3  & $0.5$ & $2.63$ & 24  & 1  & $0.01$ & $35.58$ & $83.74$ & 10 & $0.5$ & $2.58$ \\
27 & 24 & $0.5$ & $2.44$ & 49  & 1  & $0.01$ & $15.75$ & $128.99$ & 10 & $0.5$ & $2.55$ \\
& & &                    & 90  & 10 & $0.01$ & $6.8$ \\
& & &                    & 190 & 10 & $0.05$ & $5.25$ \\
\cmidrule{1-12} \\
\cmidrule{1-12}
\multicolumn{4}{c}{\textbf{KAR2}} & \multicolumn{4}{c}{\textbf{Reverse Diffusion}} & \multicolumn{4}{c}{\textbf{Euler-Maruyama}} \\
\cmidrule(lr){1-4} \cmidrule(lr){5-8} \cmidrule(lr){9-12}
$n_{den}$ & $n_{skip}$ & $\eta$ & FID & $n_{den}$ & $n_{skip}$ & $\eta$ & FID & $n_{den}$ & $n_{skip}$ & $\eta$ & FID \\
\cmidrule{1-12}
9  & 1  & $0.5$ & $6.99$ & 9   & 1  & $0.01$ & $300.18$ & 9  & 1  & $0.01$ & $306.98$ \\
23 & 2  & $0.5$ & $3.17$ & 24  & 1  & $0.01$ & $94.98$  & 24 & 1  & $0.01$ & $47.83$ \\
31 & 20 & $0.5$ & $2.64$ & 49  & 1  & $0.01$ & $22.1$   & 49 & 1  & $0.02$ & $12.73$ \\
51 & 50 & $0.5$ & $2.19$ & 90  & 10 & $0.01$ & $8.61$   & 90 & 10 & $0.05$ & $4.79$ \\
& & &                    & 190 & 10 & $0.1$  & $4.45$   & 190 & 10 & $0.1$ & $2.69$ \\
& & &                    & 370 & 30 & $1.0$  & $2.64$   & 370 & 30 & $1.0$ & $2.22$ \\
& & &                    & 570 & 30 & $1.0$  & $2.14$ \\
\bottomrule
\end{tabular}}
\caption{DLG hyper-parameters and FID for integrators in Figure \ref{fig:cifar10_acc}.}
\label{tab:cifar10_param}
\end{table*}

\newpage

\textbf{CelebA-HQ-256 image generation.} We use a pre-trained NCSN++ (cont.) score model provided by \citet{song2021}. For the baseline methods, we use the recommended settings. For DLG, the chain was initialized by generating samples with the stochastic integrator of \citet{karras2022} using 37 NFE, adding Gaussian noise of variance $0.25$, and running 70 iterations of Langevin-Gibbs. Table \ref{tab:celeba_param}  lists the hyper-parameters for DLG and the corresponding FID used to produce Figure \ref{fig:celeba_acc}.

\begin{table*}[h!]
\centering
\resizebox{0.8\textwidth}{!}{
\begin{tabular}{c c c c c c c c c c c c}
\toprule
\multicolumn{4}{c}{\textbf{KAR1}} & \multicolumn{4}{c}{\textbf{Probability Flow}} & \multicolumn{4}{c}{\textbf{RK45}} \\
\cmidrule(lr){1-4} \cmidrule(lr){5-8} \cmidrule(lr){9-12}
$n_{den}$ & $n_{skip}$ & $\eta$ & FID & $n_{den}$ & $n_{skip}$ & $\eta$ & FID & $n_{den}$ & $n_{skip}$ & $\eta$ & FID \\
\cmidrule{1-12}
17 & 4  & $0.8$ & $14.11$ & 140 & 5 & $0.25$ & $28.37$ & $125.95$ & 4 & $0.8$ & $15.29$ \\
\cmidrule{1-12} \\
\cmidrule{1-12}
\multicolumn{4}{c}{\textbf{KAR2}} & \multicolumn{4}{c}{\textbf{Reverse Diffusion}} & \multicolumn{4}{c}{\textbf{Euler-Maruyama}} \\
\cmidrule(lr){1-4} \cmidrule(lr){5-8} \cmidrule(lr){9-12}
$n_{den}$ & $n_{skip}$ & $\eta$ & FID & $n_{den}$ & $n_{skip}$ & $\eta$ & FID & $n_{den}$ & $n_{skip}$ & $\eta$ & FID \\
\cmidrule{1-12}
37 & 10 & $0.8$ & $8.78$ & 100 & 25 & $0.1$ & $26.43$ & 100 & 4 & $0.8$ & $23.19$ \\
\bottomrule
\end{tabular}}
\caption{DLG hyper-parameters and FID for deterministic samplers in Figure \ref{fig:celeba_acc}.}
\label{tab:celeba_param}
\end{table*}

\textbf{Achieving SOTA.} On CIFAR10, we use KAR1 settings of Table \ref{tab:cifar10_param}. On CelebA-HQ-256, we use $n_{den} = 131$, $n_{skip} = 27$, $\eta = 4.0$, which achieves $6.99$ FID.

\section{Additional Samples}

\subsection{DLG Chain Visualization}

\begin{figure}[h]
\centering
\includegraphics[width=1.0\linewidth]{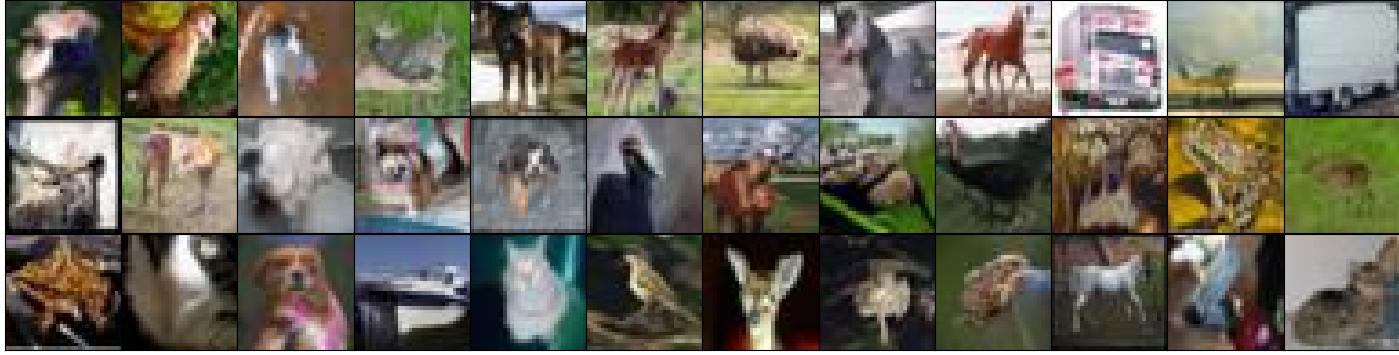}
\caption{Visualization of a DLG chain on CIFAR10.}
\label{fig:cifar10_mcmc}
\end{figure}

\begin{figure}[h]
\centering
\includegraphics[width=1.0\linewidth]{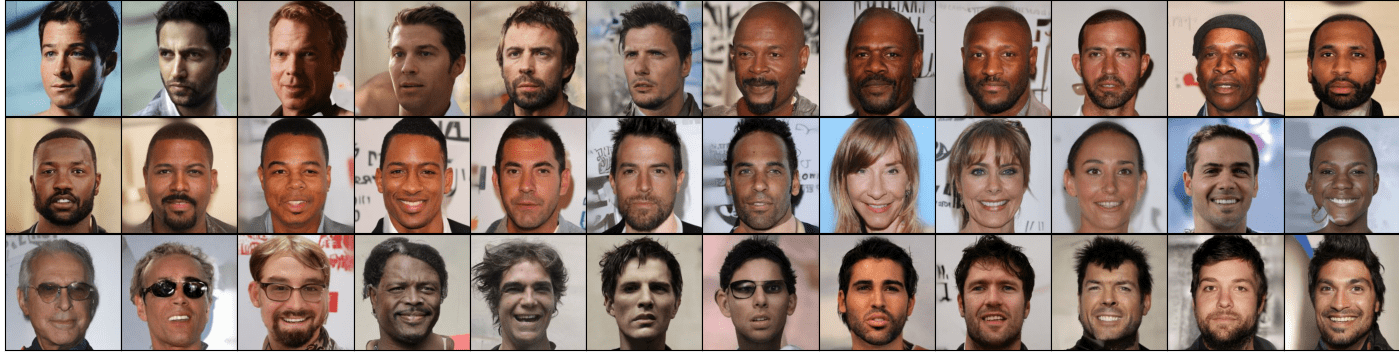}
\caption{Visualization of a DLG chain on CelebA-HQ-256.}
\label{fig:celeba_mcmc}
\end{figure}

Figures \ref{fig:cifar10_mcmc} and \ref{fig:celeba_mcmc} each show a DLG chain on CIFAR10 and CelebA-HQ-256, respectively. The chain progresses from left to right, from right end to left end of row below. On CIFAR10, we can see that the chain visits diverse classes. On CelebA-HQ-256, we can see that the chain transitions between diverse attributes such as gender, hair color, skin color, glasses, facial expression, posture, etc.

\subsection{Additional Unconditional Samples}

\begin{figure}[t]
\centering
\includegraphics[width=0.49\linewidth]{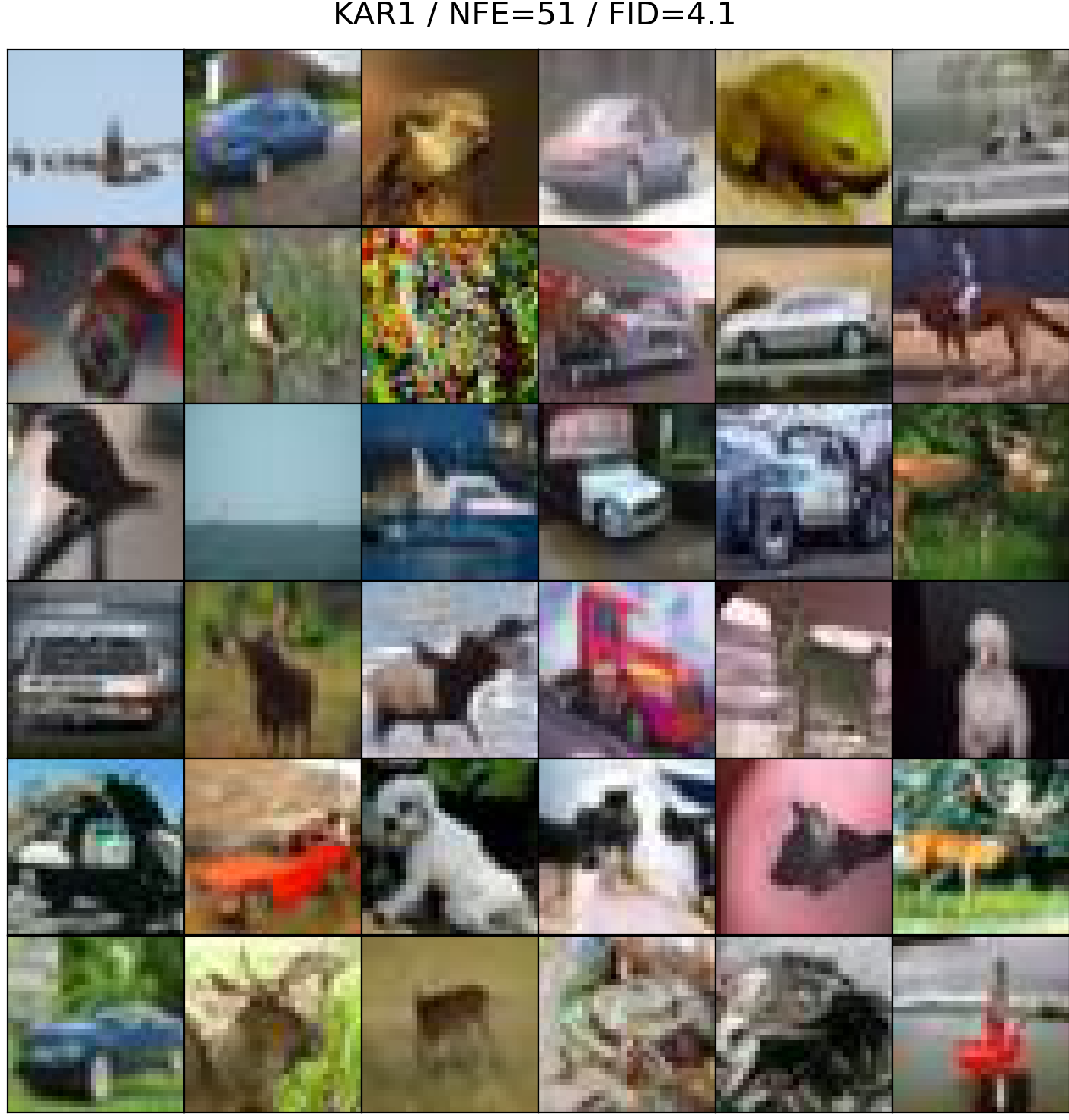}
\hfill
\includegraphics[width=0.49\linewidth]{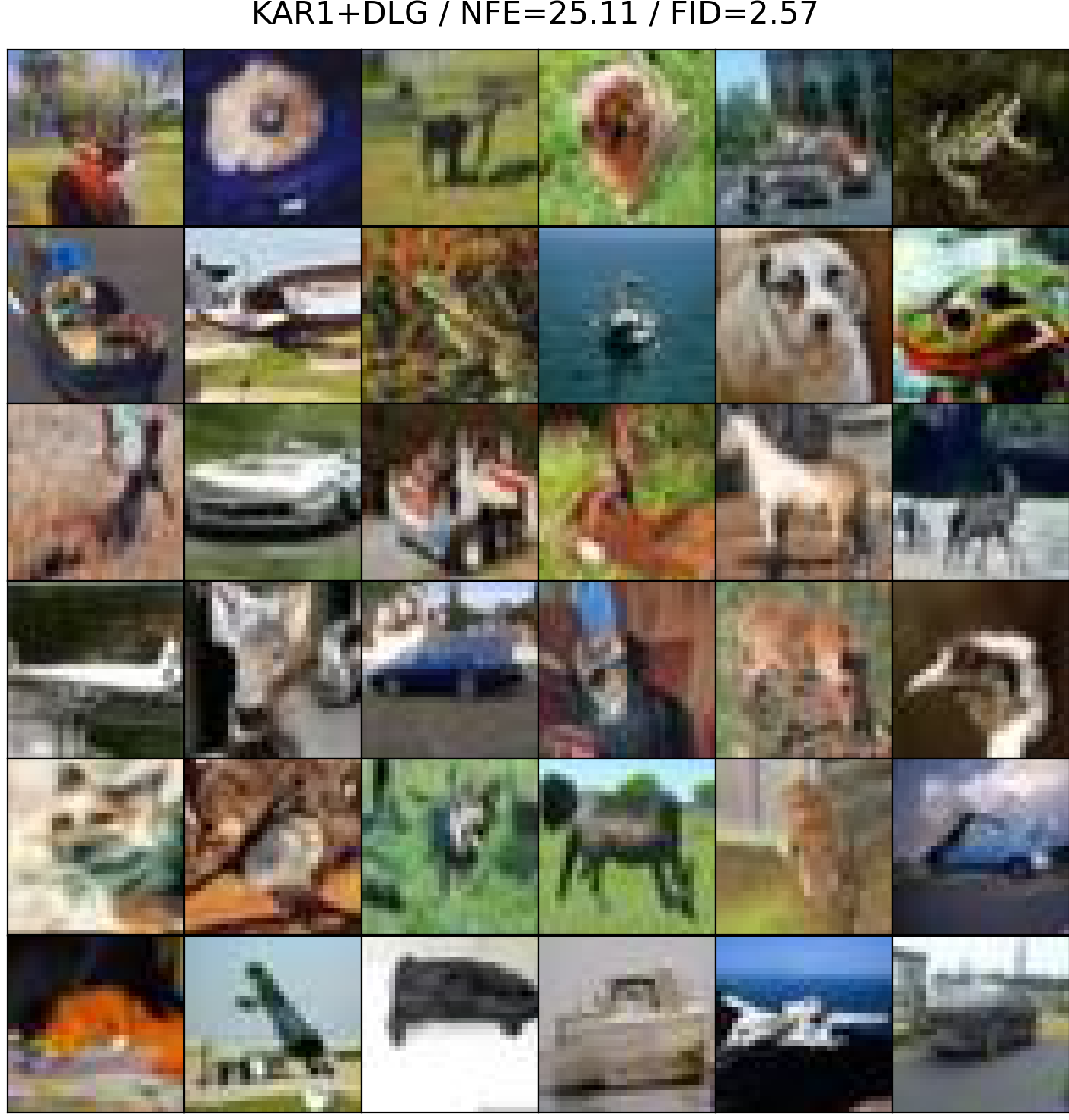} \\ [1em]
\includegraphics[width=0.49\linewidth]{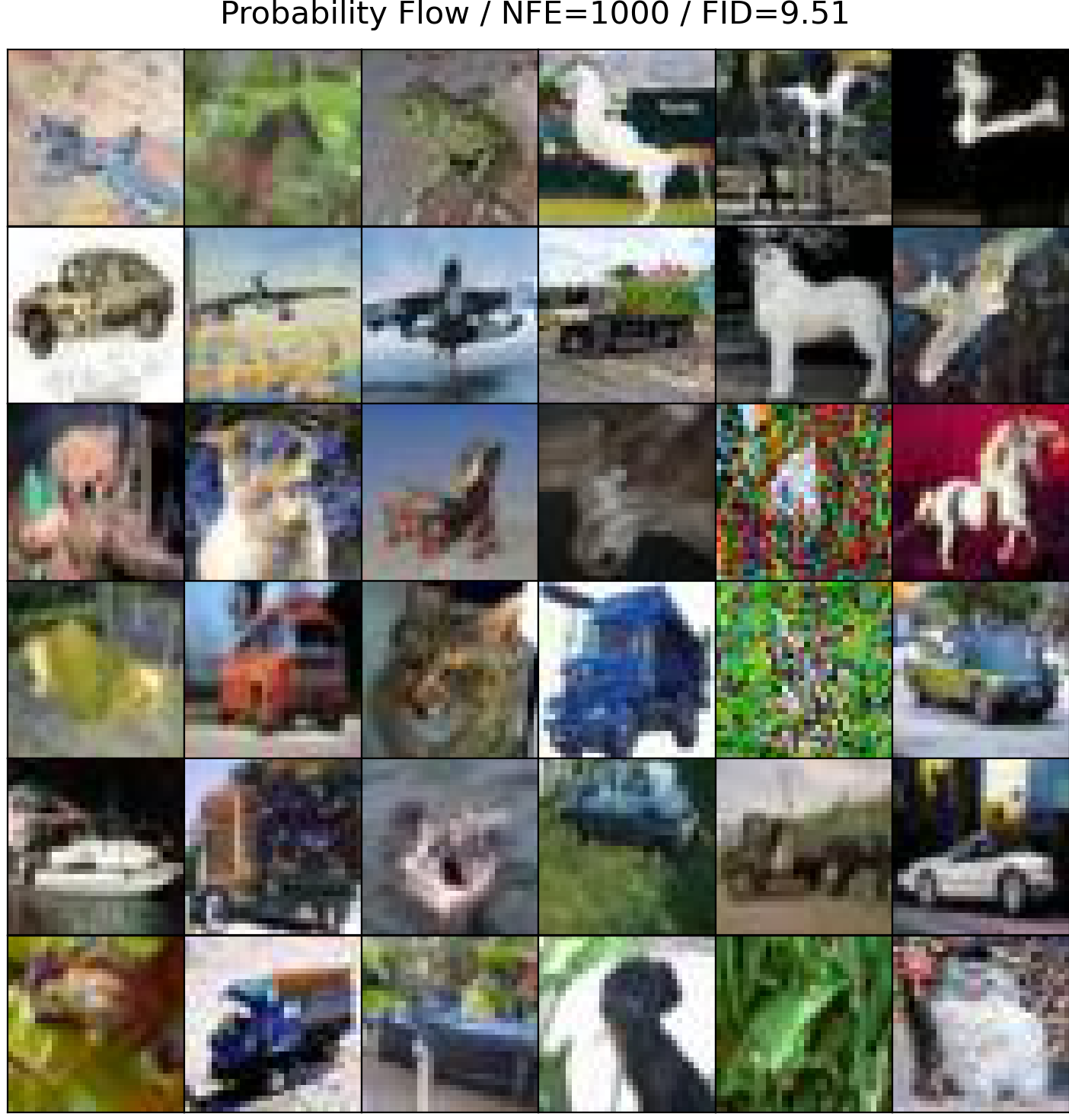}
\hfill
\includegraphics[width=0.49\linewidth]{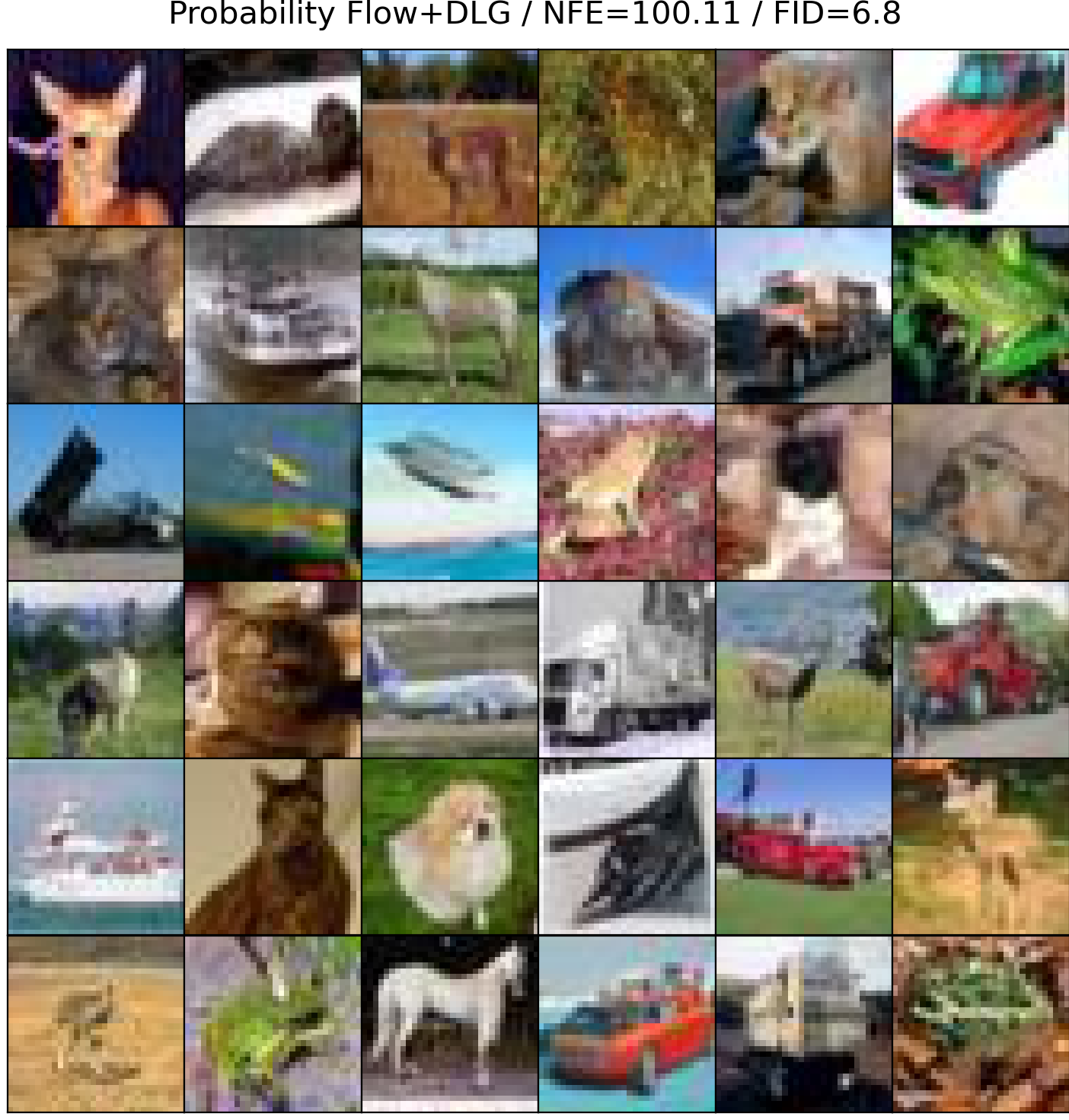} \\ [1em]
\includegraphics[width=0.49\linewidth]{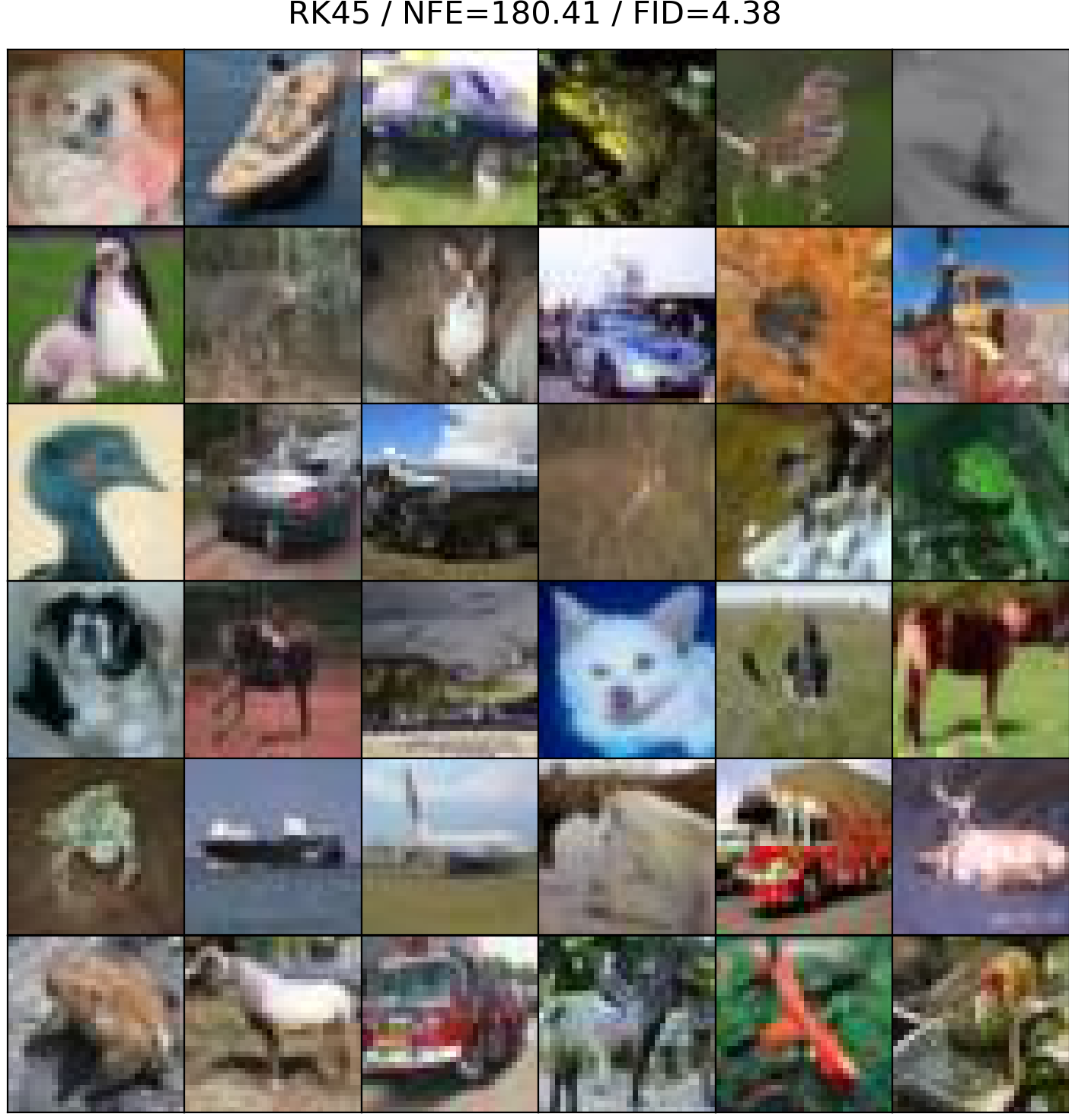}
\hfill
\includegraphics[width=0.49\linewidth]{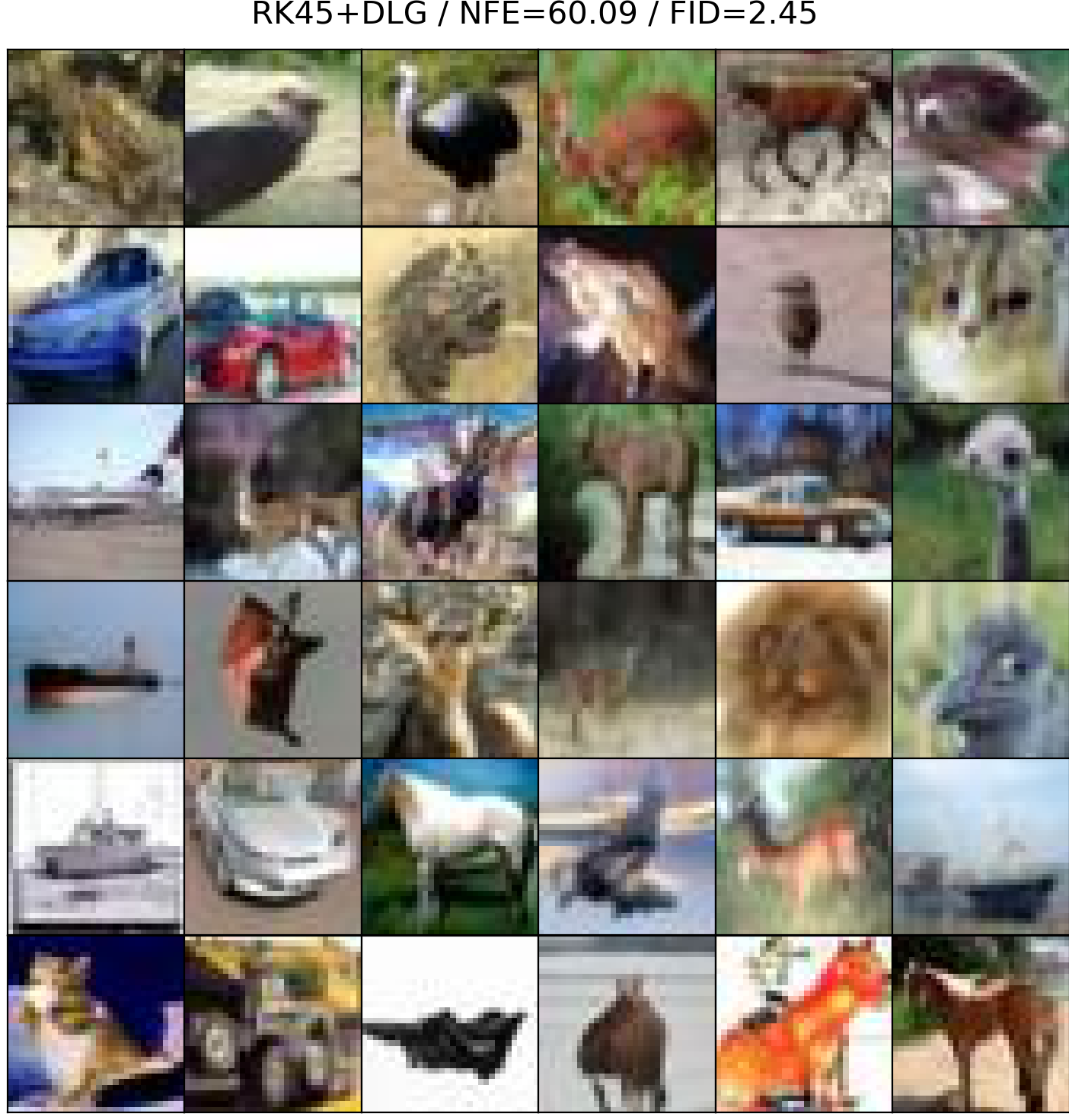}
\caption{Additional non-cherry-picked samples for deterministic integrators on CIFAR10 without (left col.) and with (right col.) DLG.}
\label{fig:cifar10_det_samples}
\end{figure}

\begin{figure}[t]
\centering
\includegraphics[width=0.49\linewidth]{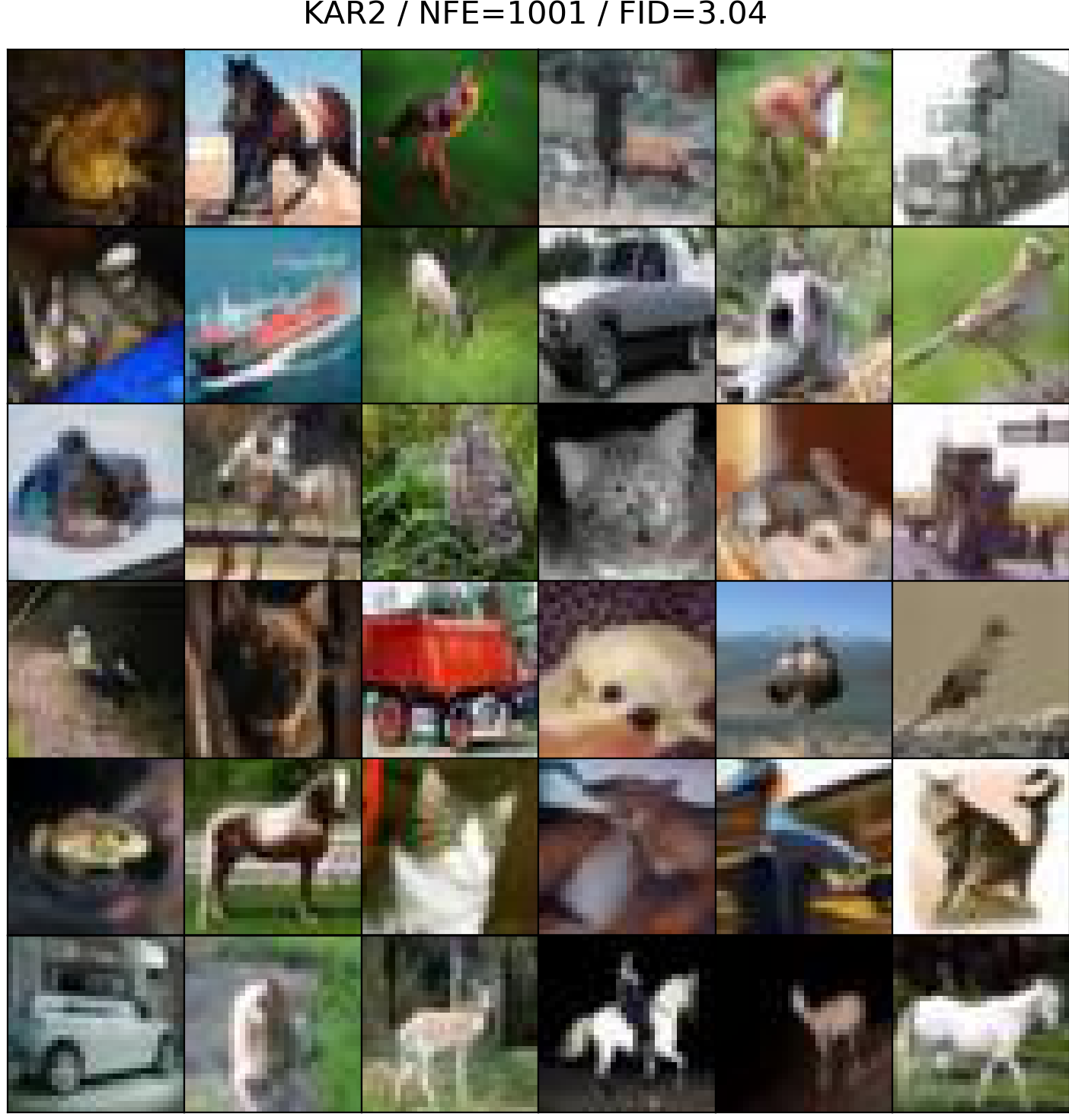}
\hfill
\includegraphics[width=0.49\linewidth]{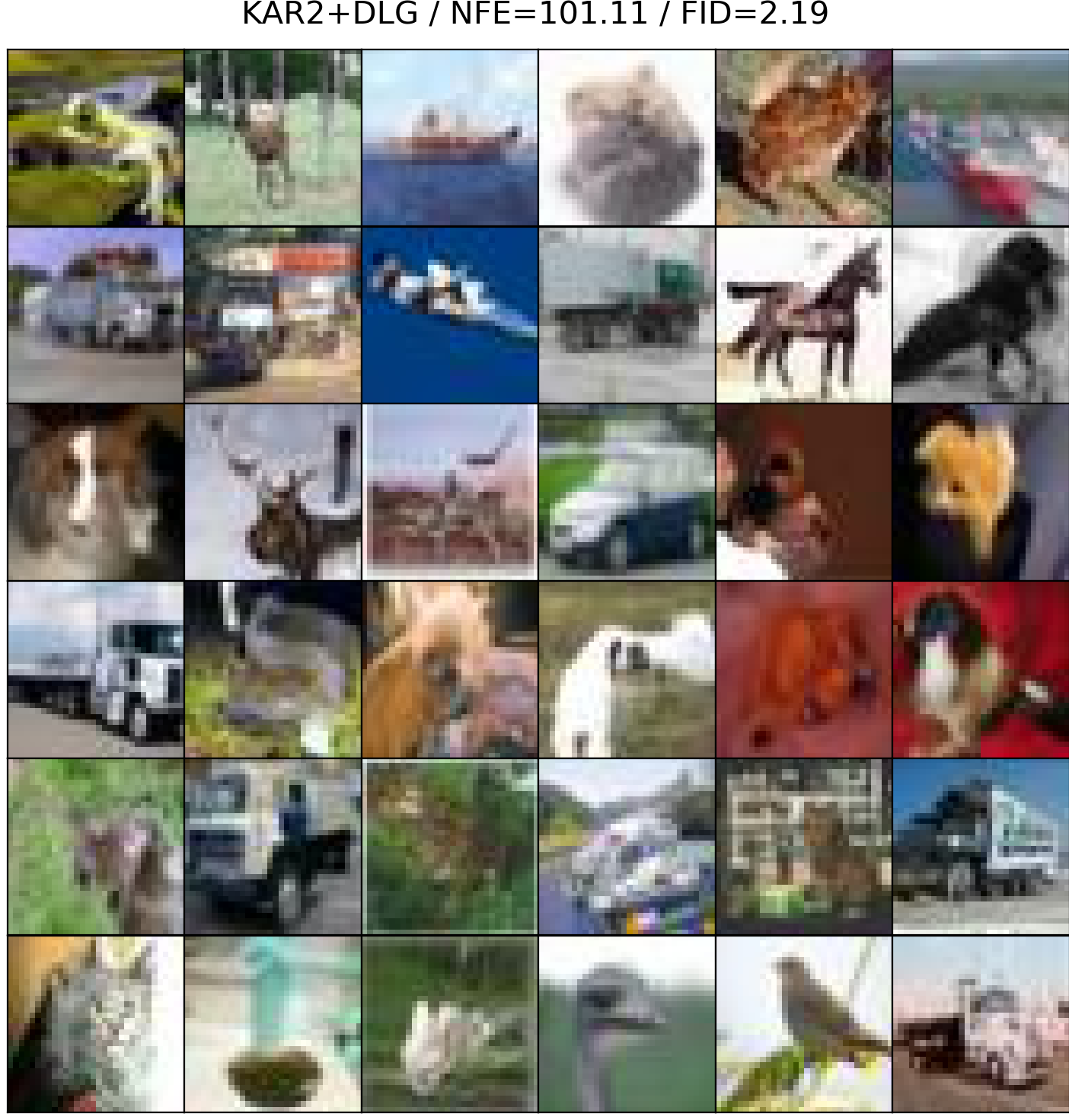} \\ [1em]
\includegraphics[width=0.49\linewidth]{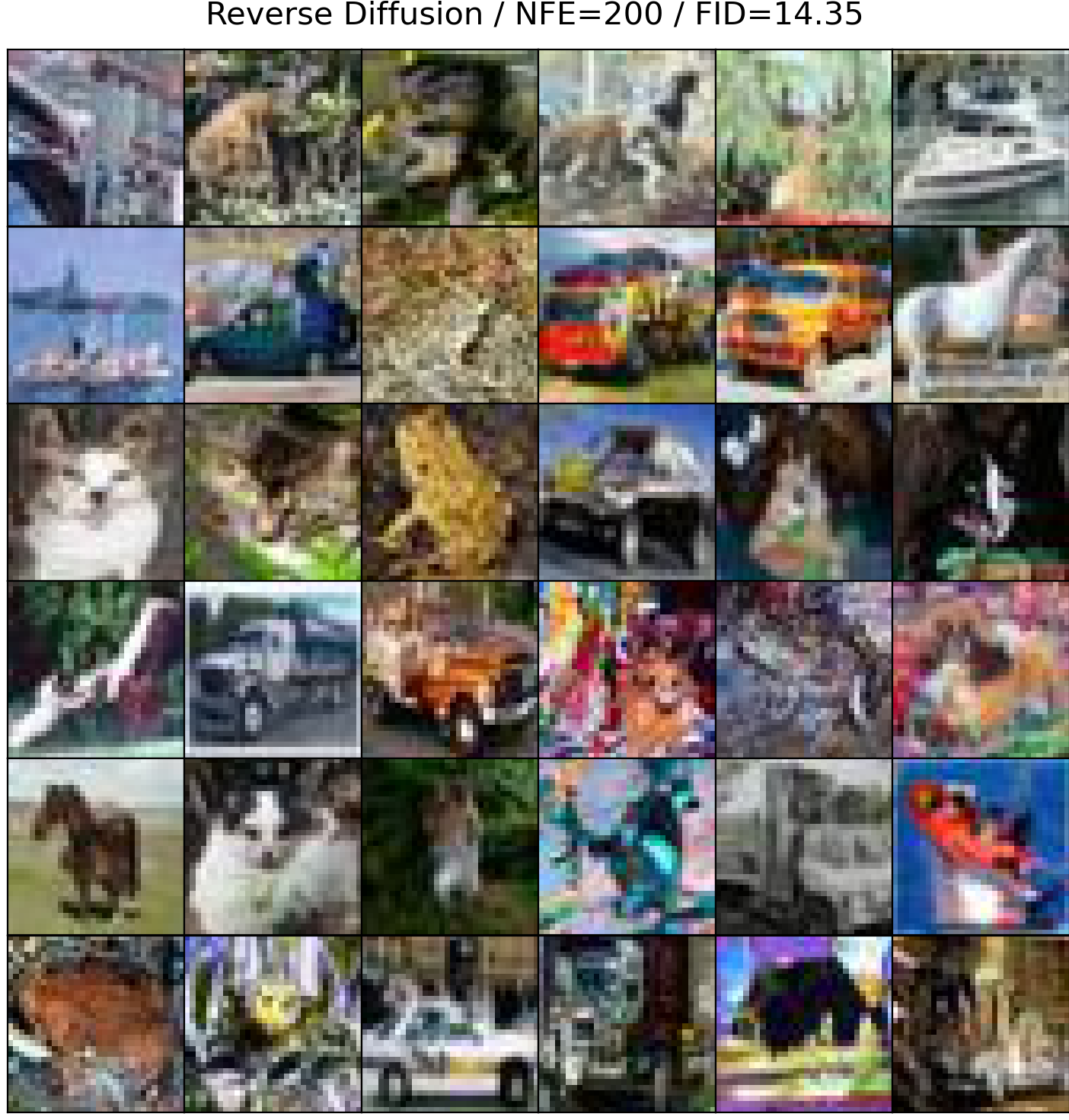}
\hfill
\includegraphics[width=0.49\linewidth]{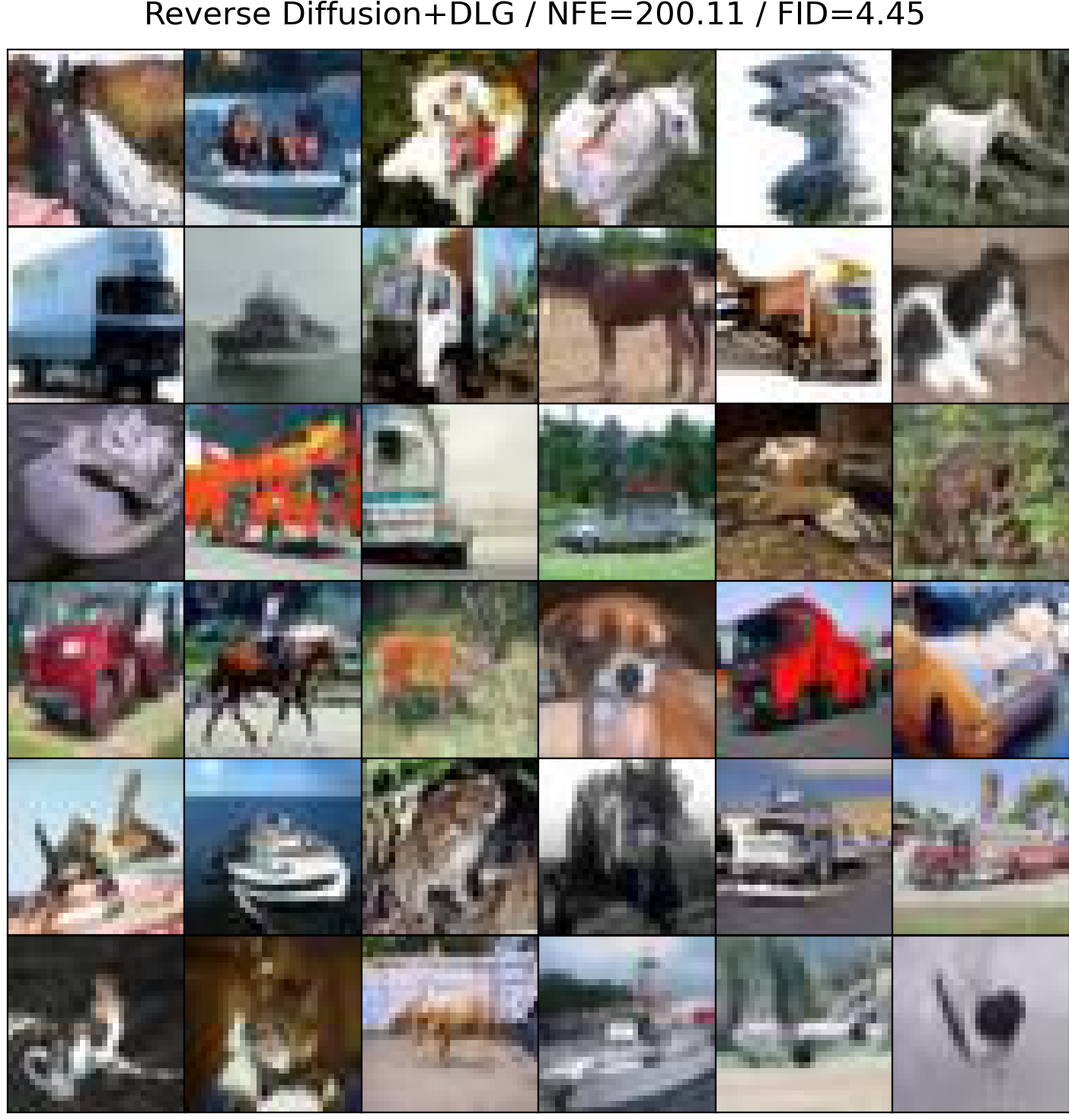} \\ [1em]
\includegraphics[width=0.49\linewidth]{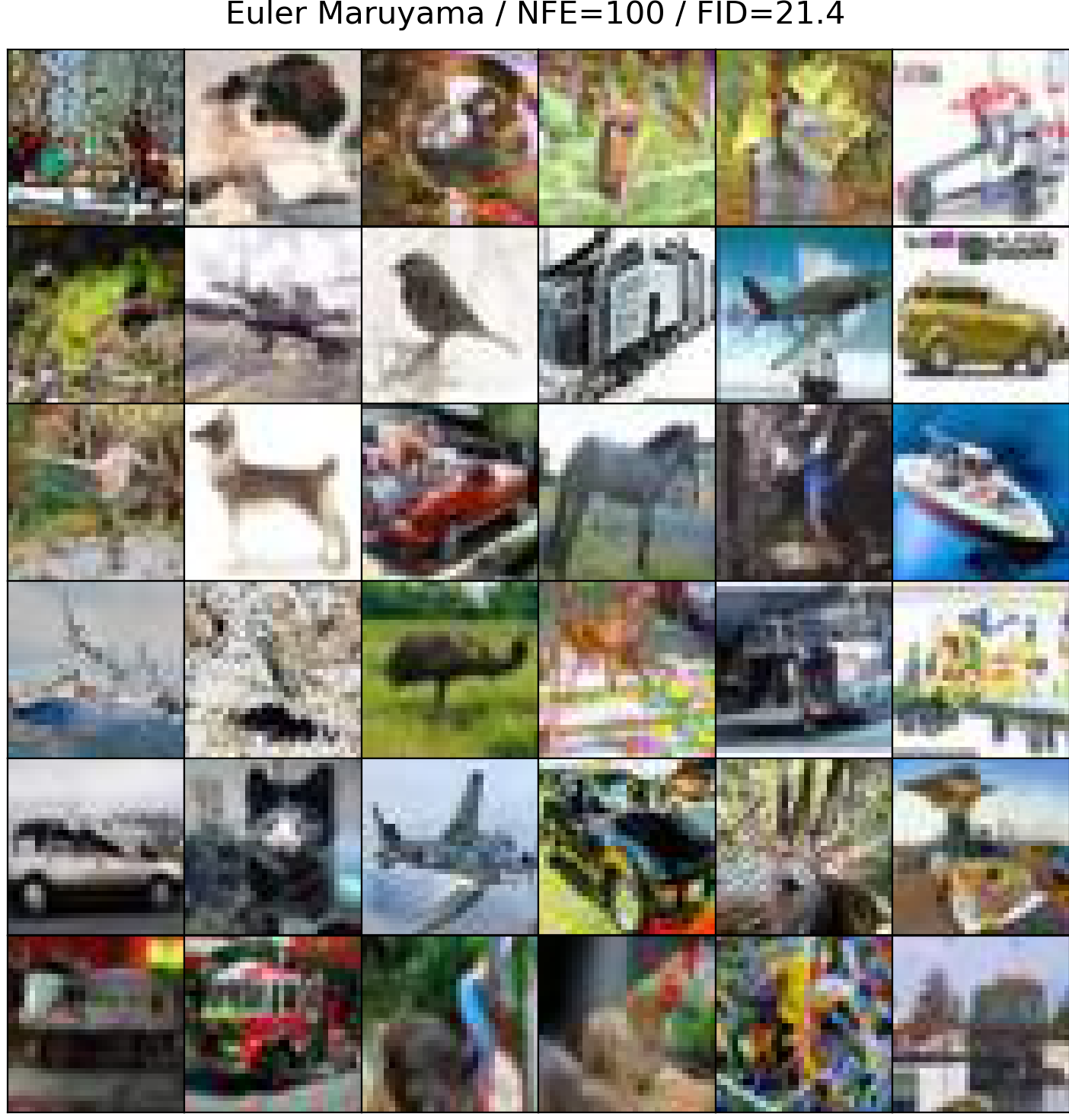}
\hfill
\includegraphics[width=0.49\linewidth]{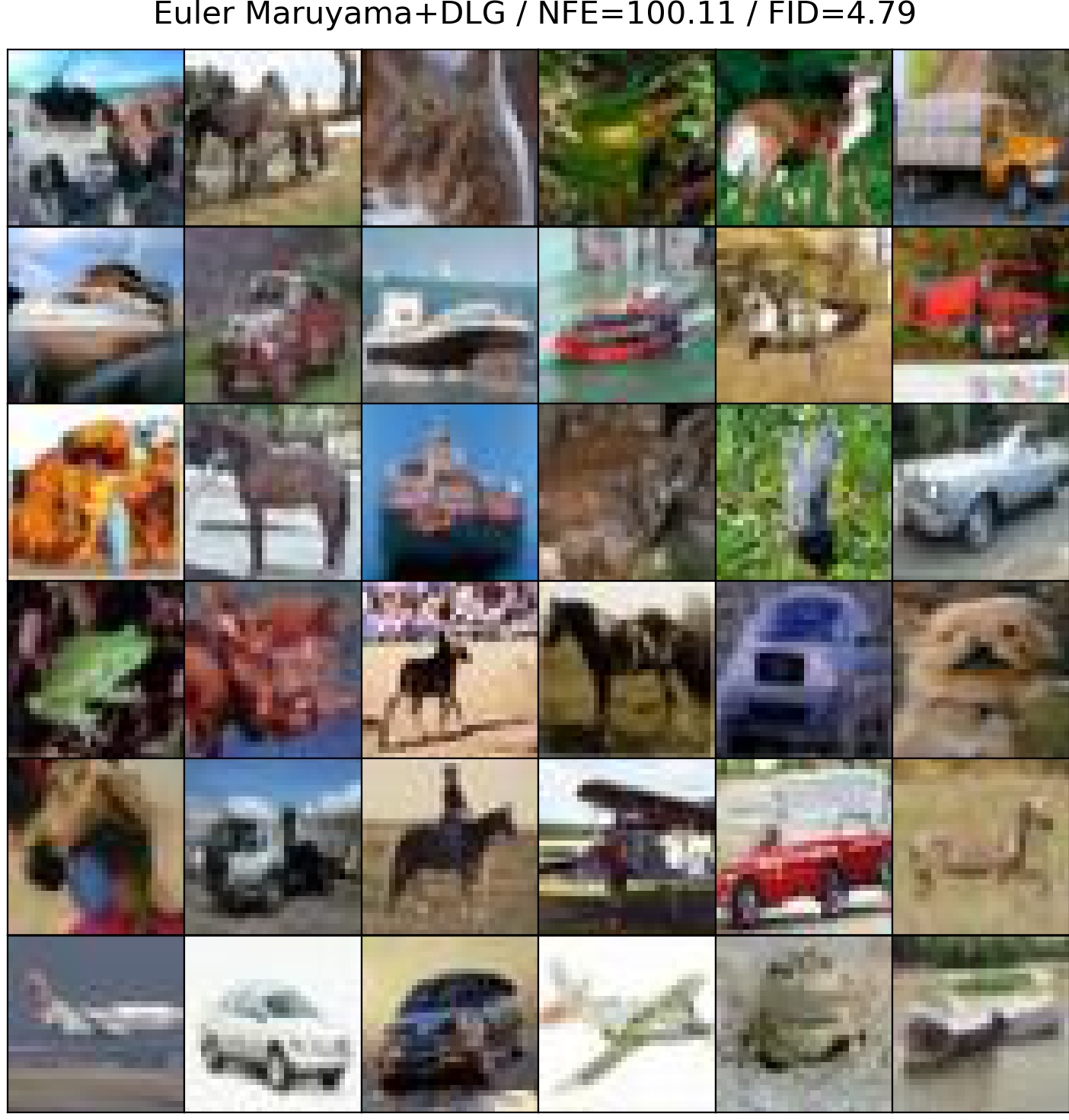}
\caption{Additional non-cherry-picked samples for stochastic integrators on CIFAR10 without (left col.) and with (right col.) DLG.}
\label{fig:cifar10_stoc_samples}
\end{figure}

\begin{figure}[t]
\centering
\includegraphics[width=0.49\linewidth]{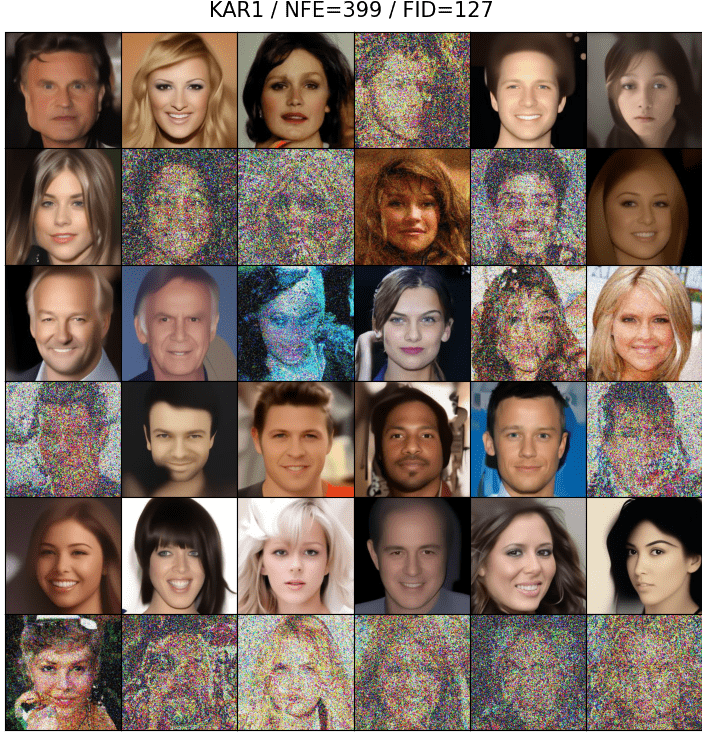}
\hfill
\includegraphics[width=0.49\linewidth]{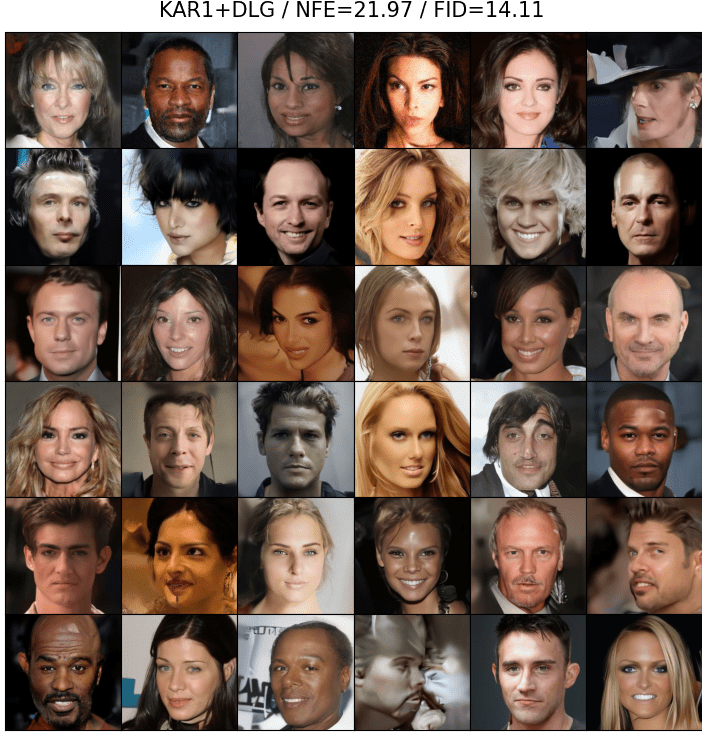} \\ [1em]
\includegraphics[width=0.49\linewidth]{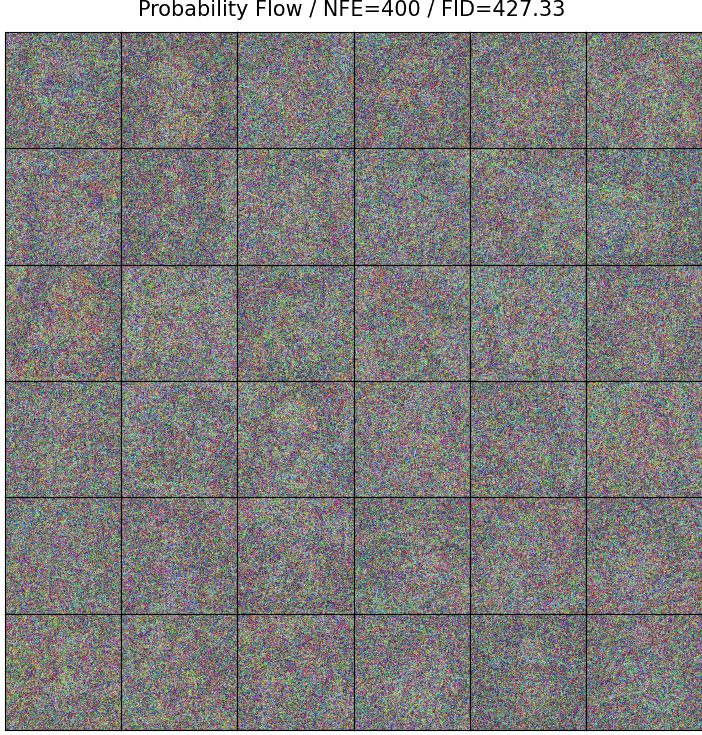}
\hfill
\includegraphics[width=0.49\linewidth]{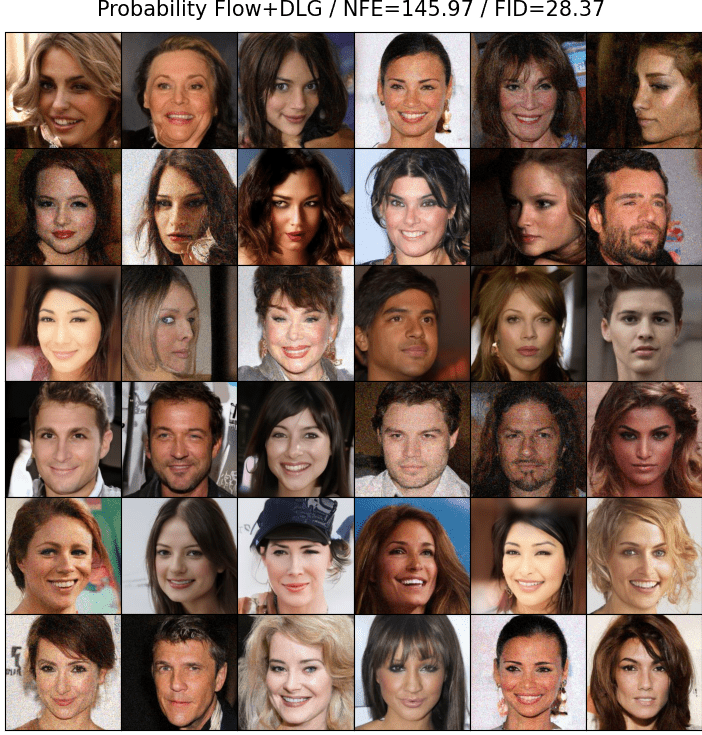} \\ [1em]
\includegraphics[width=0.49\linewidth]{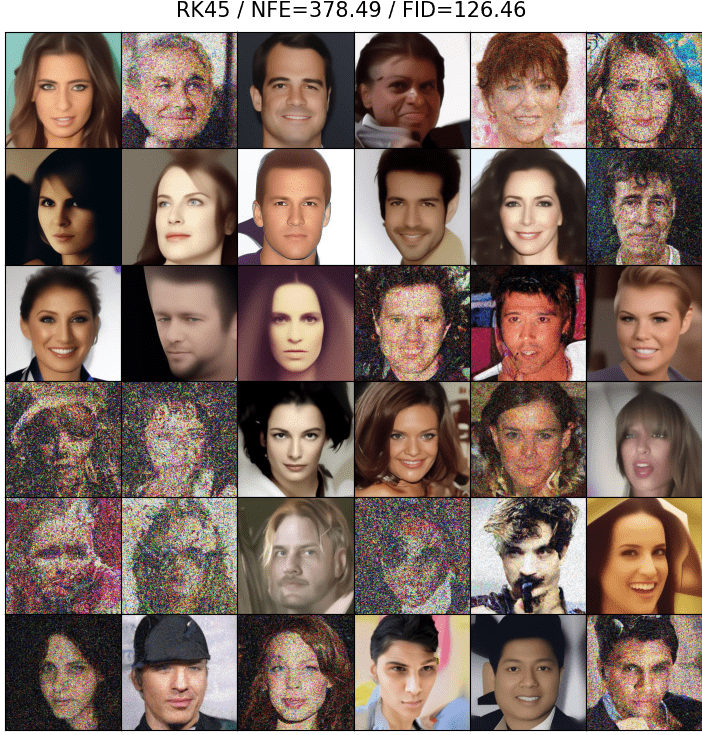}
\hfill
\includegraphics[width=0.49\linewidth]{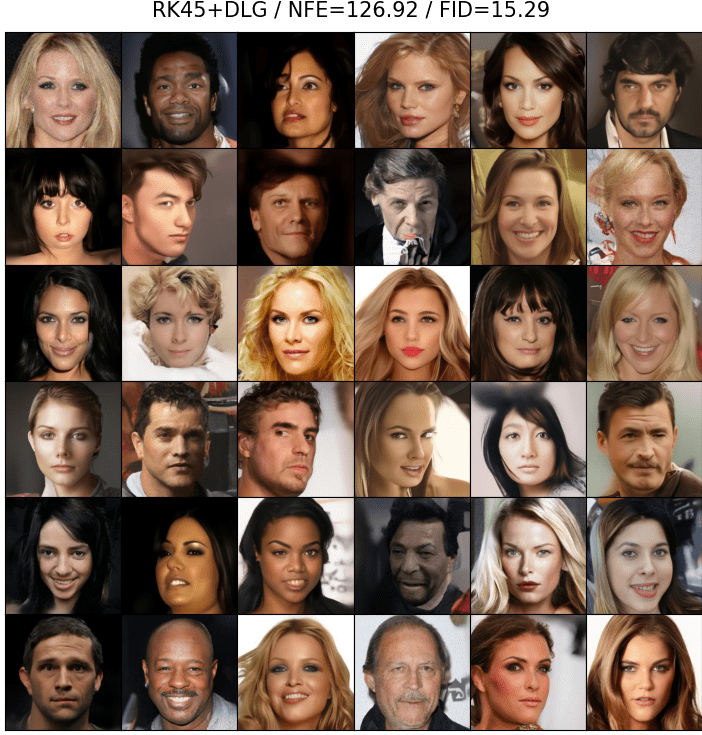}
\caption{Additional non-cherry-picked samples for deterministic integrators on CelebA-HQ-256 without (left col.) and with (right col.) DLG.}
\label{fig:celeba_det_samples}
\end{figure}

\begin{figure}[t]
\centering
\includegraphics[width=0.49\linewidth]{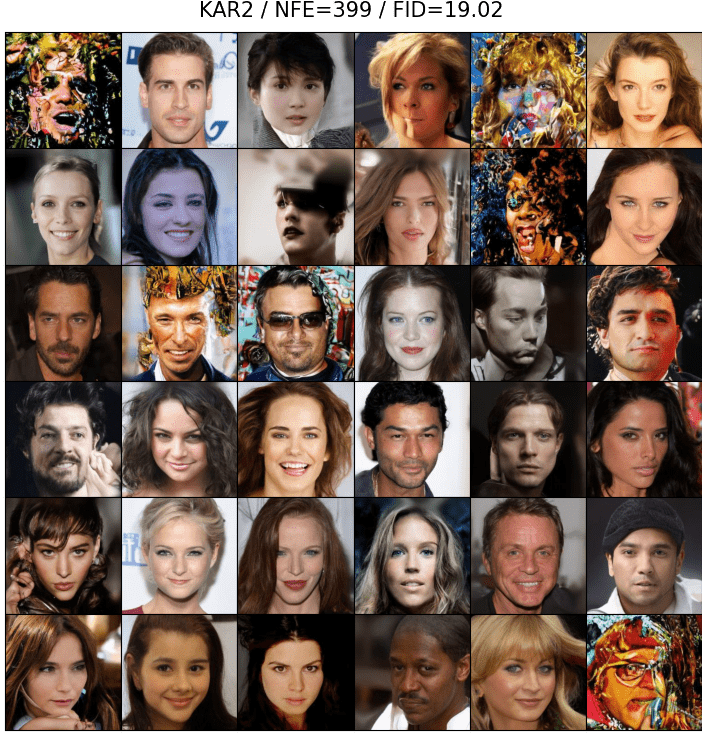}
\hfill
\includegraphics[width=0.49\linewidth]{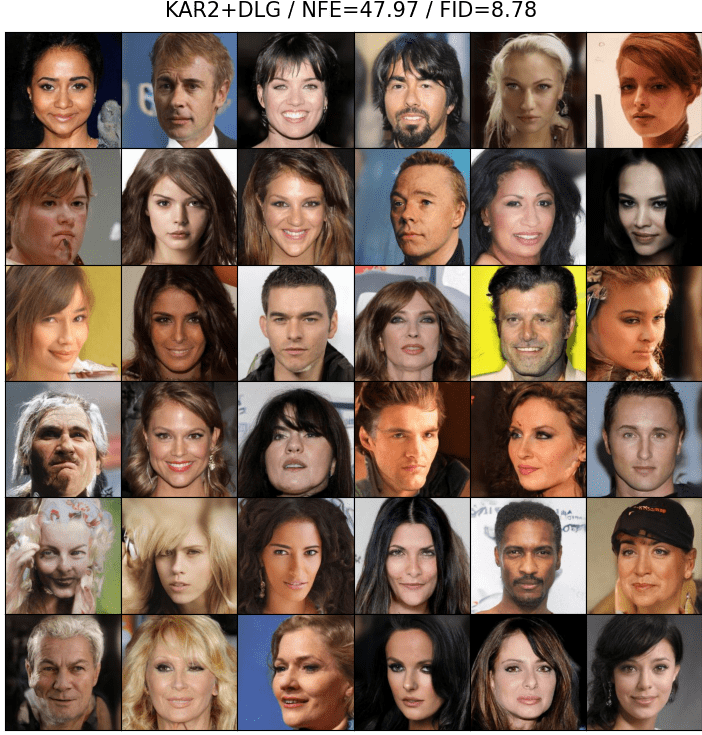} \\ [1em]
\includegraphics[width=0.49\linewidth]{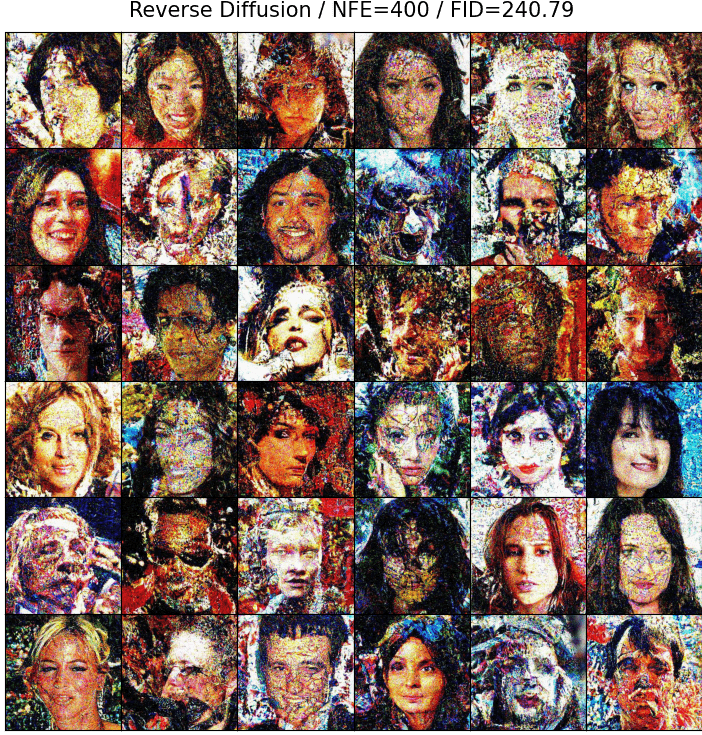}
\hfill
\includegraphics[width=0.49\linewidth]{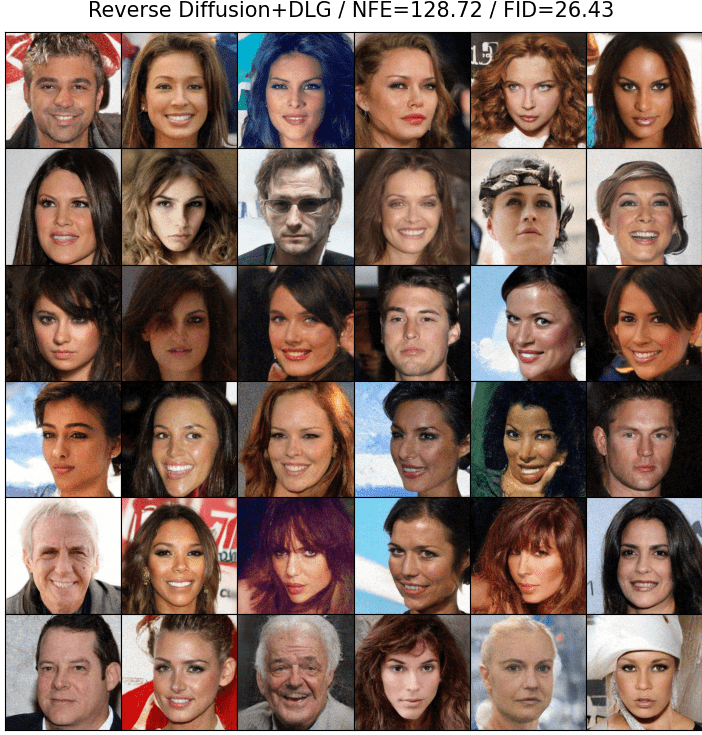} \\ [1em]
\includegraphics[width=0.49\linewidth]{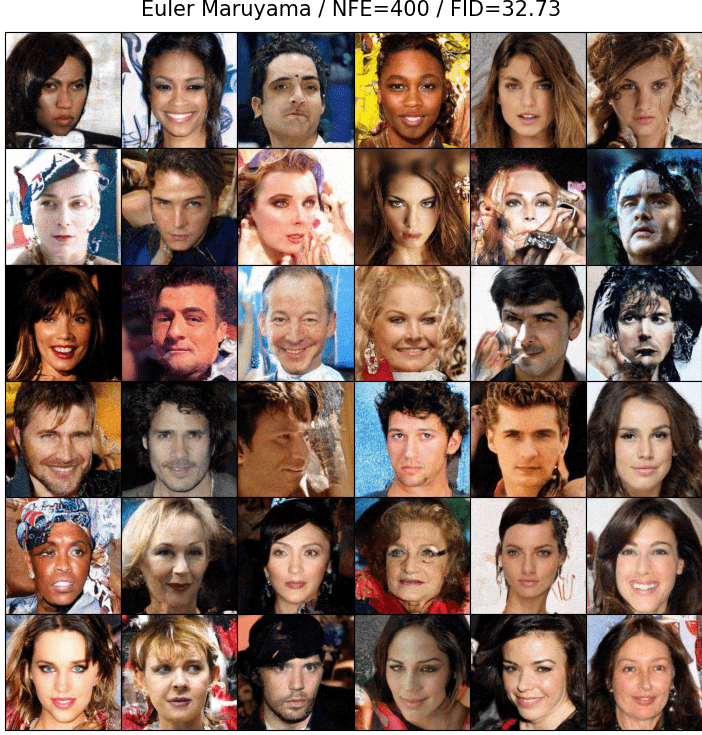}
\hfill
\includegraphics[width=0.49\linewidth]{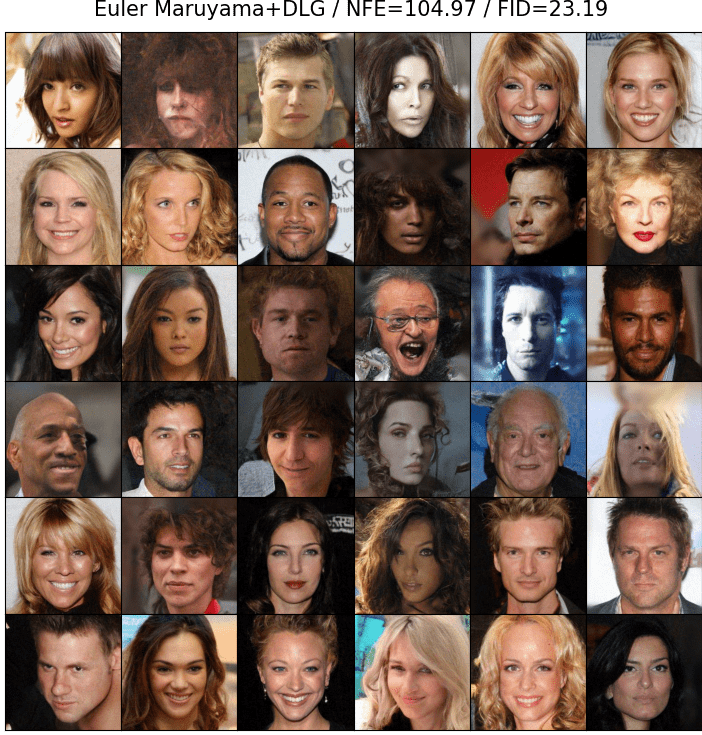}
\caption{Additional non-cherry-picked samples for stochastic integrators on CelebA-HQ-256 without (left col.) and with (right col.) DLG.}
\label{fig:celeba_stoc_samples}
\end{figure}

In Figures \ref{fig:cifar10_det_samples} and \ref{fig:cifar10_stoc_samples}, we show additional samples for CIFAR10 without and with DLG. In Figures \ref{fig:celeba_det_samples} and \ref{fig:celeba_stoc_samples}, we show additional samples for CelebA-HQ-256 without and with DLG.

\newpage

\newpage

\newpage

\newpage

\end{document}